%% file: acl_latex.tex
\documentclass[11pt]{article}

\usepackage{authblk}
\usepackage[final]{acl}

\usepackage{times}
\usepackage{latexsym}

\usepackage{physics,amsmath}
\usepackage{cleveref}
\usepackage{enumitem}
\usepackage{booktabs}

\usepackage{textcomp, xspace}
\usepackage{pbox}
\usepackage{array}
\usepackage{multirow}
\usepackage{xcolor}
\usepackage[normalem]{ulem}
\useunder{\uline}{\ul}{}
\usepackage{rotating}
\usepackage{array}
\usepackage{listings}
\usepackage{adjustbox}

\usepackage{amssymb}
\usepackage{arydshln}
\usepackage[most]{tcolorbox}
\usepackage{csquotes}
\usepackage{fvextra}
\usepackage{fontawesome5}
\usepackage{graphicx}
\usepackage{makecell}

\newcommand{\doublewings}{%
  \reflectbox{\faFeather}\faFeather%
}

\usepackage[T1]{fontenc}
\usepackage[utf8]{inputenc}

\usepackage{microtype}

\usepackage{inconsolata}

\usepackage{graphicx}

\newcommand{\green}[1]{\colorbox{green!45}{#1}}
\newcommand{\red}[1]{\colorbox{magenta!45}{#1}}
\newcommand{\blue}[1]{\colorbox{cyan!45}{#1}}
\newcommand{\purple}[1]{\colorbox{violet!45}{#1}}

\newcommand{\model}{\textsc{TSUBASA}\xspace}

\title{\model\doublewings: Improving Long-Horizon Personalization \\via Evolving Memory and Self-Learning with Context Distillation}

\author{\textbf{Xinliang Frederick Zhang}}
\author{\textbf{Lu Wang}}

\affil{Computer Science and Engineering, University of Michigan, Ann Arbor, MI} \affil{\{\texttt{xlfzhang,wangluxy\}@umich.edu}}

\begin{document}
\maketitle
\begin{abstract}
Personalized large language models (PLLMs) have garnered significant attention for their ability to align outputs with individual's needs and preferences. 
However, they still struggle with long-horizon tasks, such as tracking a user's extensive history of conversations or activities. 
Existing memory mechanisms often fail to capture evolving behaviors, and RAG paradigms are trapped by a \textit{quality-efficiency tradeoff}. Meanwhile, parametric adaptation is bottlenecked by \textit{train-inference gap} due to the scarcity of labeled data. 
To enhance the long-horizon capabilities of PLLMs, 
we introduce \model, a two-pronged approach designed to improve \textit{memory writing} via dynamic memory evolution, and \textit{memory reading} via self-learning with a context distillation objective to internalize user experiences. 
Extensive evaluations on long-horizon benchmarks using the \textsc{Qwen}-3 model family (4B to 32B) validate the effectiveness of \model, surpassing competitive memory-augmented systems that rely primarily on memory writing, such as Mem0 and Memory-R1. 
Our analyses further confirms that \model breaks the quality-efficiency barrier to achieve Pareto improvements, delivering robust, high-fidelity personalization with a reduced token budget.
\end{abstract}

\input{1_Intro}

\input{2_Related}

\input{3_Prelim}

\input{4_Method}

\input{5_Experiment}

\input{6_Results}

\input{7_Conclusion}

\section*{Acknowledgments}
This work was supported in part by the Air Force Office of Scientific Research under grant FA9550- 22-1-0099,  and computational resources and services provided by Advanced Research Computing (ARC), a division of Information and Technology Services (ITS) at the University of Michigan, Ann Arbor.

\input{Limitation}

\setcounter{table}{0}
\setcounter{figure}{0}
\renewcommand{\thefigure}{A\arabic{figure}}
\renewcommand{\thetable}{A\arabic{table}}

% \clearpage

\appendix

\input{Appendix}

\end{document}

%% file: 1_Intro.tex
\section{Introduction}

\begin{figure*}[t]
    \centering
    \vspace{-14mm}
    \includegraphics[width=0.9\textwidth]{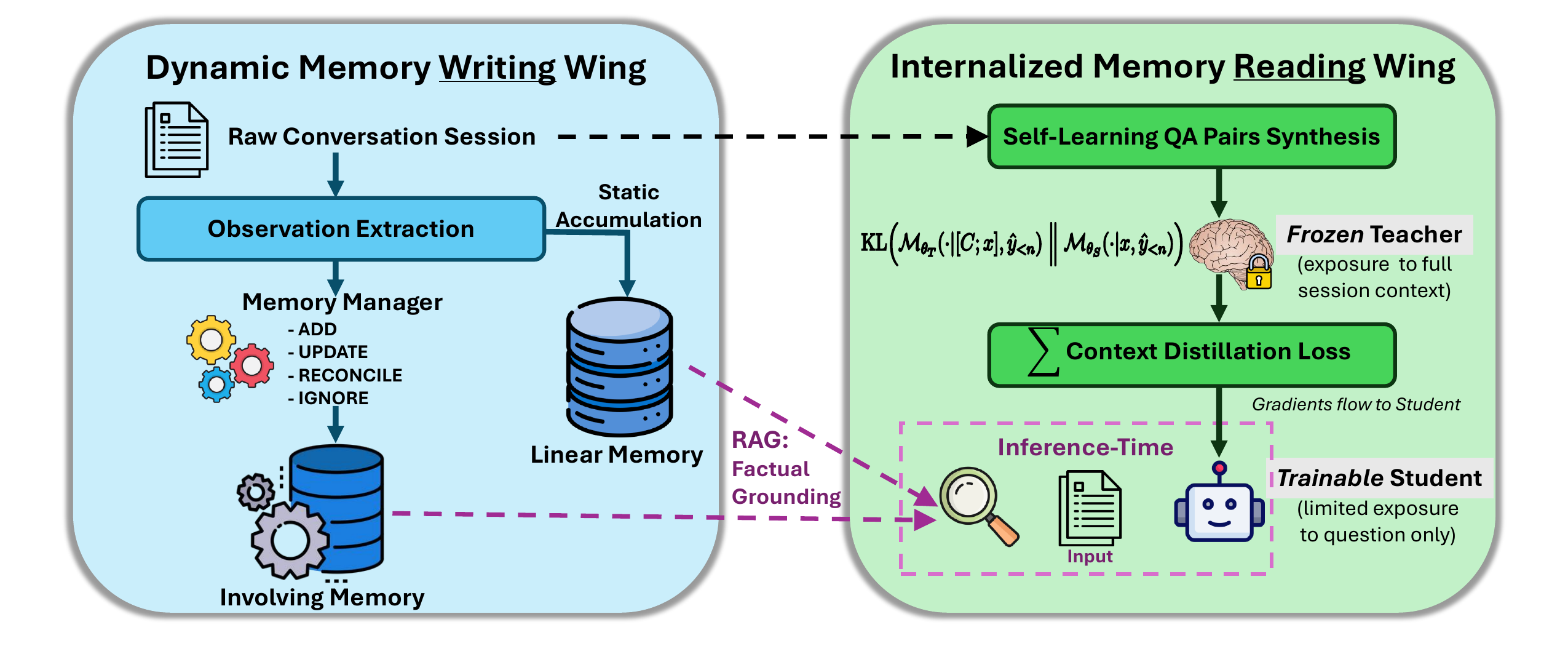}
    \vspace{-4mm}
    \caption{Overview of \model framework, built on two synergistic wings. \blue{\textbf{Dynamic memory writing}}  applies structured algorithmic evolution based on high-density observations distilled from raw utterances (\Cref{sec:mem_writing}). 
    \green{\textbf{Internalized memory reading}} adopts the self-learning pipeline and applies context distillation objective on synthetic data (\Cref{sec:mem_reading}). During \purple{Inference}, observations are retrieved for factual grounding.} 
    \label{fig:overall_architecture}
    \vspace{-6mm}
\end{figure*}

In recent years, large language models (LLMs)
have evolved from passive text generators into versatile reasoning engines capable of analyzing diverse data modalities and performing deep reasoning~\citep{chen2025towards, comanici2025gemini}.
However, despite their vast parametric knowledge, generic LLMs often follow a \textit{one-size-fits-all} approach that fails to cater to individuals' nuanced, idiosyncratic requirements~\citep{liu2025survey, qiu-etal-2025-measuring, duran2025beyond}.
To bridge this gap, 
personalized LLMs (PLLMs) utilize user context and parametric adaptation to align outputs with individualized needs and specific user personas~\citep{salemi-etal-2024-lamp, zhang-etal-2025-prime}.

Recent advancements in PLLMs 
span prompt engineering~\citep{park2023generative, DBLP:journals/corr/abs-2310-20081}, retrieval-augmented generation~\citep[RAG; ][]{salemi-etal-2024-lamp, mysore-etal-2024-pearl}, lightweight parameterization~\citep{DBLP:journals/corr/abs-2404-03565, magister-etal-2025-way}, and hybrid \textit{dual-memory} models that mirror human episodic and semantic memory~\citep{zhang-etal-2025-prime}.
Despite these efforts, existing methods fundamentally struggle with \textbf{long-horizon personalization} due to their inability to \textit{grasp continuously evolving user preferences and a critical lack of labeled user data for parametric memory updates}.
Concretely, standard episodic memory systems often act as linear-growing vector databases~\citep{lewis2020retrieval, madaan-etal-2022-memory}, failing on ``reflective'' tasks that require implicit temporal reasoning across interactions~\citep{tan-etal-2025-membench}. While recent work explores evolving memory to address this~\citep{nan2025nemori, wang2025mem, luo2026storage}, they remain flawed: some {compromise production privacy} through cross-user training~\citep{yan2025memory}, while others rely on destructive DELETE operations~\citep{chhikara2025mem0} that erase historical context and obscure the cause-and-effect behind user evolution. 
Furthermore, internalizing memory via parameterization~\citep{DBLP:journals/corr/abs-2404-03565, DBLP:journals/corr/abs-2411-13405} requires abundant, hard-to-acquire labeled data.

To advance the memory mechanisms and unlock deep personalization for long-horizon tasks, PLLMs must achieve both \textbf{native personalized reasoning} and \textbf{access to episodic memory for factual grounding}. However, attempting to solve these high-level objectives with current architectures exposes two technical bottlenecks (\Cref{sec:prelim}). First, a \textit{train-inference gap}: standard fine-tuning on raw conversations fails to prepare parametric memory for the complex personalization tasks at evaluation. Second, a \textit{quality-efficiency tradeoff}: standard RAG paradigms correlate better response quality with larger volumes of retrieved memories, incurring prohibitive computational costs.

To this end, we introduce \model framework (\textit{lit.} \textit{``Wing''} in Japanese; \Cref{fig:overall_architecture}), standing for {\ul\textbf{T}}wo-winged per {\ul\textbf{S}}onalization  {\ul\textbf{U}}nifying  {\ul\textbf{B}}i-Directional  {\ul\textbf{A}}utonomous memory  {\ul\textbf{S}}torage and parametric  {\ul\textbf{A}}ssimilation.
We posit that an effective memory system comprises two synergistic wings (\textbf{writing} and \textbf{reading}), both essential for genuine personalization. Specifically, during \textit{dynamic memory writing} (\Cref{fig:memory_example}), core, factual \textbf{observations} are extracted from raw conversations, and then a memory manager \textit{autonomously} updates the memory store to prevent redundancy and resolve conflicts. During \textit{internalized memory reading}, we employ a self-learning pipeline to tackle the lack of labeled user-specific data and the knowledge gap inherent in naive autoregressive training. We augment this with a teacher-student context distillation objective~\citep[\cref{eq}; ][]{snell2022learning} on \textit{synthetic user data}, enabling the teacher to distill rich ``dark knowledge'' about the user directly into the student.

We conduct extensive empirical evaluations on LoCoMo~\citep{maharana-etal-2024-evaluating} and LongMemEval~\citep{DBLP:conf/iclr/WuWYZCY25} to assess PLLMs in long-horizon conversational settings. Our results validate the high information density of extracted observations, the efficacy of internalized memory reading, the necessity of explicit RAG-backed grounding, and the power of our evolving memory algorithm especially when the model size scales.
Remarkably, in terms of F1 results, \model-PRO demonstrates substantial gains: it outperforms widely used systems like Mem0~\citep{chhikara2025mem0} by nearly $50\%$ overall and surpasses the prior SOTA, Memory-R1~\citep{yan2025memory}, by $4.9\%$, without breaking data privacy constraints. Thanks to our self-learning pipeline, it drives $17\%-49\%$ improvements on single- and multi-hop QA tasks.
Furthermore, our analysis confirms that \model breaks the quality-efficiency barrier to achieve strict \textit{Pareto improvements}.

In summary, our contributions are three-fold:
\begin{itemize} [leftmargin=2em,itemsep=0em,topsep=1pt,parsep=1pt,partopsep=1pt]
\item We identify two critical bottlenecks trapping existing PLLMs: the train-inference gap in semantic parameterization and the quality-efficiency tradeoff for episodic retrieval.
\item We propose \model, a novel framework for long-horizon tasks featuring two synergistic wings: dynamic memory writing and internalized memory reading.
\item Extensive experiments showcase the effectiveness of \model, highlighting its ability to achieve strict Pareto improvements while preserving user privacy.
\end{itemize}

%% file: 2_Related.tex
\section{Related Work}

\subsection{LLM Personalization}
\label{sec:llm_personalization}
The pursuit of adapting models to human nuances has driven extensive research \citep{DBLP:journals/corr/abs-2402-09660}. Early efforts evolved from explicit modeling of user profiles and demographic attributes \citep[e.g., age, gender; ][]{DBLP:conf/ercimdl/AmatoS99, fischer2001user, kim2013personality, DBLP:conf/recsys/GaoHBLLZ13, DBLP:conf/chi/GouZY14} to latent-factor models and collaborative filtering that map user-item interactions into low-dimensional embeddings \citep{DBLP:journals/computer/KorenBV09, DBLP:journals/tkde/JiangCWZY14, he2017neural, FAREED2023100495}. While Transformer~\citep{transformer} enabled learnable user embeddings \citep{DBLP:conf/aaai/QiuWG021, deng-etal-2023-annotate}, these approaches struggle to leverage \textit{unstructured} text or generalize across diverse tasks, necessitating frameworks capable of deeper semantic alignment with raw textual histories.

With the advent of LLMs, initial efforts relied on prompt engineering for static personas \citep{park2023generative, tang-etal-2024-medagents, tseng-etal-2024-two} or profile summaries \citep{DBLP:journals/corr/abs-2310-20081}, often yielding superficial, surface-level stylistic mimicry. To improve robustness, systems began incorporating entire interaction histories \citep{DBLP:journals/corr/abs-2304-10149, DBLP:journals/corr/abs-2306-11114, DBLP:journals/corr/abs-2305-06474}. To overcome context window limits and ``lost-in-the-middle'' issue \citep{liu-etal-2024-lost}, retrievers like BM25 \citep{DBLP:journals/ftir/RobertsonZ09} and FAISS \citep{douze2024faiss} were adopted to only keep relevant items~\citep{madaan-etal-2022-memory, salemi-etal-2024-lamp, mysore-etal-2024-pearl}. Yet, noisy retrieval still restricts fine-grained personalization. As such, recent studies internalize historical engagement directly into model weights via lightweight adapters \citep{tan-etal-2024-personalized, tan-etal-2024-democratizing, DBLP:journals/corr/abs-2404-03565, magister-etal-2025-way}. Inspired by cognitive theories \citep{tulving1972episodic, tulving1985many}, unified frameworks like PRIME combine RAG-based discrete episodic memory with parameterized semantic memory to further enhance capabilities \citep{zhang-etal-2025-prime}.

Despite this progress, existing unified, dual-memory architectures face two critical limitations. First, discrete episodic memory struggles to capture the \textit{evolving} nature of user preferences, which is integral to long-horizon tasks \citep{wei2025evo, chhikara2025mem0}. Second, parametric memory updating typically demand \textit{labeled} user data \citep{zhang-etal-2024-llm-based, salemi-etal-2024-lamp}, which is frequently unavailable in practice. Thus, our goal is to advance the dual-memory architecture to address these specific challenges in long-horizon setting.

\subsection{Memory Mechanisms for Personalization}
\label{sec:memory4pllm}
Decades of psychological research have established core human memory components, ranging from short-term to durable long-term memory, which is further divided into episodic and semantic memory structures~\citep{AtkinsonShiffrin1968, tulving1972episodic, tulving1985many}. Imitating these mechanisms is central to LLM personalization, acting as an architectural cornerstone to overcome the inherent \textit{statelessness} of plain LLMs~\citep{zhang2024survey, wu2025human, du2025rethinking}. Traditional methods rely on brute-force caching of raw conversational logs, but this \textit{static accumulation} fails catastrophically on ``reflective'' tasks requiring implicit reasoning across interactions~\citep{tan-etal-2025-membench}. Consequently, the field has witnessed a rapid shift towards autonomously evolving memory systems that utilize reflection and abstraction to capture dynamic user preferences~\citep{wang2024survey, luo2026storage}. These techniques range from static summarization with decay of older, less relevant memories~\citep{zhong2024memorybank, salama-etal-2025-meminsight} to truly dynamic approaches executing autonomous, incremental updates~\citep{kim2025pre, nan2025nemori, xu2025mem, rasmussen2025zep, wang2025mem}.

A critical divergence in memory evolution algorithms is \textit{how} memory updates (e.g., ADD, UPDATE, DELETE, NOOP) are executed. To perform these actions, frameworks like Memory-R1~\citep{yan2025memory} and Mem-$\alpha$~\citep{wang2025mem} train dedicated managers via PPO~\citep{schulman2017proximal} or GRPO~\citep{shao2024deepseekmath}, while Mem0~\citep{chhikara2025mem0} prompts proprietary models like GPT-4~\citep{achiam2023gpt}. However, these approaches introduce severe bottlenecks: RL pipelines are notoriously unstable and demand domain-specific reward models, whereas proprietary APIs incur prohibitive costs and latency. This makes them unsuitable for portable, on-device deployments with strict computational limits~\citep{qin2024enabling, wang2025never}.
Furthermore, existing frameworks mishandle evolving preferences by using a destructive DELETE operation~\citep{yan2025memory, chhikara2025mem0}, which erases historical context and obscures the cause-and-effect of users' mindset shifts. To address this, we introduce a novel RECONCILE action and incorporate temporal signals to build coherent narratives, preserving the absolute integrity of the user timeline.

%% file: 3_Prelim.tex
\section{Prelim: Challenges of Memory Usage}
\label{sec:prelim}

To motivate our  solution, we first rigorously examine the performance bottlenecks inherent in prevailing approaches. Specifically, we evaluate the LOCOMO dataset~\citep{maharana-etal-2024-evaluating} using the best-performing, portable LLMs~\citep[\textsc{Qwen3}; ][]{yang2025qwen3}, to highlight two fundamental challenges in \textit{how answer agents utilize memory}.

\paragraph{Memory Structure and the Train-Inference Gap.}
The first fundamental challenge concerns how to \textit{instantiate memory}. An intuitive baseline to internalize user context is to naively fine-tune the LLM using an autoregressive next-token prediction objective over the raw conversational history (i.e., \textit{semantic memory usage}). However, empirical results (\Cref{tbl:prelim_results}) demonstrate a counter-intuitive reality: this baseline actually underperforms an untrained LLM augmented with RAG (i.e., \textit{episodic memory usage}). 
This discrepancy stems from a non-trivial \textit{train-inference gap}. During naive autoregressive training, the model optimizes for the rote memorization of conversations, including irrelevant chatter. Yet, at inference time, complex personalization tasks~\citep[e.g., multi-hop QA; ][]{maharana-etal-2024-evaluating} demand deeper reasoning to piece together temporally scattered facts. 
Meanwhile, relying solely on RAG to retrieve raw utterances is equally \textit{flawed}. Raw dialogues contain massive semantic noise and repeated information, are temporally fragmented, and act merely as external prompts that do not contribute to genuine, internalized personalization.

\paragraph{The Quality-Efficiency Tradeoff.} 
The second major challenge arises from the inherent tradeoff between inference efficiency and response quality. In the standard RAG paradigm, efficiency is strictly bottlenecked by the volume of retrieved memories~\citep{madaan-etal-2022-memory, salemi-etal-2024-lamp, mysore-etal-2024-pearl, zhong2024memorybank}. As shown in \Cref{tbl:prelim_results}, both \textsc{Qwen3}-8B and \textsc{Qwen3}-14B exhibit the same pattern: full \textit{session}\footnote{A session is an ordered sequence of interactions between two participants within a short, temporally bounded interval.} context exposure consistently yields the best results but consumes an average of $2,158.5$ tokens per query. Conversely, restricting the context window to only the single most relevant memory piece dramatically improves inference speed ($29.4$ tokens) but significantly degrades accuracy. 
In fact, real-world applications demand PLLMs that are both quality and efficient, this tradeoff highlights an imperative need for architectural advancement: the length of the historical context must be explicitly decoupled from inference-time compute, while the LLM should still retain the full access to the historical knowledge without the associated latency costs.

\input{Tables/prelim}

\noindent Next, we describe our proposed solution, \model, where the dynamic memory writing module addresses the \textit{raw utterance usage} issue~(i.e., \textit{low information density, high redundancy, and lack of temporal awareness}), and the internalized memory reading module tackles both \textit{train-inference gap} and \textit{quality-efficiency tradeoff}.

%% file: Tables/prelim.tex
\begin{table}[t]
\centering
% \vspace{-2mm}
\resizebox{0.95\linewidth}{!}{%
\begin{tabular}{lrrrrrrr} \toprule
\multirow{2}{*}{Method} & \multicolumn{3}{c}{\textsc{Qwen3}-8B}                                                      & \multicolumn{3}{c}{\textsc{Qwen3}-14B}                                                     & \multicolumn{1}{c}{\multirow{2}{*}{\makecell{Input Len.\\(tokens)}}} \\ \cmidrule(lr){2-4} \cmidrule(lr){5-7}
                        & {F1} & {BLEU-1} & {Rouge-L} & {F1} & {BLEU-1} & {Rouge-L} & \multicolumn{1}{c}{}                           \\  \midrule
RAG (k=1)               & 21.26                  & 17.06                      & 20.82                       & 24.29                  & 20.00                      & 23.78                       & 29.4                                           \\
RAG (k=3)               & 28.10                  & 22.53                      & 27.30                       & 31.38                  & 26.02                      & 30.59                       & 75.4                                           \\
Train                   & 6.07                   & 4.32                       & 5.73                        & 6.59                   & 4.96                       & 6.24                        & 5.6                                            \\
Session                 & 41.22                  & 33.64                      & 39.61                       & 44.26                  & 36.40                      & 42.87  & 2{,}158.5  \\ \bottomrule                     
\end{tabular}
}
\vspace{-2mm}
\caption{Preliminary study reveals \textit{train-inference gap} and \textit{quality-efficiency tradeoff} as discussed in~\Cref{sec:prelim}. }
\vspace{-6mm}
\label{tbl:prelim_results}
\end{table}

%% file: 4_Method.tex
\section{Methodology: \model Framework}

\model (\Cref{fig:overall_architecture}) is constructed upon two synergistic wings: \textit{dynamic memory writing} (\Cref{sec:mem_writing}) via structured algorithmic evolution, and \textit{internalized memory reading} (\Cref{sec:mem_reading}) through self-learning with context distillation objective.

\subsection{Dynamic Memory Writing: From Raw Storage to Evolving Observation}
\label{sec:mem_writing}

To address the aforementioned \textit{raw utterance} issue, our memory writing mechanism first extracts core, factual \textbf{observations} from raw conversations before storage, analogous to the hierarchical event extraction processes~\citep{zhang-etal-2024-ultra}. \Cref{fig:prompt_extraction} in appendix details the instruction used to extract these salient observations from a session. This process \textbf{improves the information density}, \textbf{reduces intra-session redundancy} and significantly cuts the memory usage for each incoming session (e.g., store $20$ raw utterances vs. $7$ salient observations on LOCOMO dataset~\citep{maharana-etal-2024-evaluating}).

Moving beyond static accumulation, which struggles with ``reflective memory'' tasks requiring implicit reasoning across temporally-related interactions \citep{tan-etal-2025-membench}, we implement a continuous, evolutionary memory manager. To mitigate \textbf{inter-session redundancy}, new observations are not indiscriminately appended. Instead, the management agent evaluates the semantic and temporal relationship between incoming observations and the existing memory profile. Guided by evolution rules defined in a system prompt (\Cref{fig:prompt_evolution}), the manager autonomously executes one of four actions to combat linear memory growth:

\begin{itemize}[leftmargin=1em,itemsep=0em,topsep=1pt,parsep=1pt,partopsep=1pt]
    \item \textbf{ADD:} Creates a new entry for entirely novel information without duplicating existing ones.
    \item \textbf{UPDATE:} Modifies the existing memory to reflect granular details or a more recent status, avoiding unnecessary expansion.
    \item \textbf{RECONCILE:} Integrates \textit{conflicting} information into a coherent temporal narrative, preserving complete historical context while preventing retrieval confusion.
    This operation explicitly accounts for the inherently non-stationary nature of human preferences and behaviors~\citep{hoch1991time, hughes2020inferring}.
    \item \textbf{IGNORE:} Discards trivial or redundant information already captured in memory.
\end{itemize}

Notably, our RECONCILE action improves upon the destructive DELETE found in existing frameworks~\citep[i.a.,][]{yan2025memory, chhikara2025mem0}. When confronted with conflicting preferences, purging ``outdated'' facts irreparably compromises the user's engagement timeline and the cause-and-effect behind the user's evolving mindset. Instead, RECONCILE weaves conflicting information into a coherent, self-contained narrative. 
To ensure these narratives remain robust, we further augment with explicit \textbf{temporal grounding}~\citep{tan-etal-2023-towards, zhang-etal-2024-narrative, piryani2025s}. In raw conversations, temporal references are primarily relative (e.g., ``yesterday''), which quickly lose context in vector databases and corrupt retrieval quality afterwards. Thus, our dynamic writing module autonomously decodes them into \textit{absolute} calendar dates.
By implementing \textit{RECONCILE} and \textit{temporal awareness}, each memory piece now functions as a temporally grounded, conflict-free narrative (e.g., ``The user previously preferred X, but as of [Date], prefers Y''). This preserves the evolutionary trajectory of the user's behavior and provides the downstream LLM with an interpretable timeline optimized for precise retrieval.

Practically, the same underlying model performs both memory writing and reading (\Cref{sec:mem_reading}). By avoiding costly proprietary LLMs \citep{maharana-etal-2024-evaluating, chhikara2025mem0} and refraining from training on other users' data \citep{yan2025memory, wang2025mem}, this approach paves the way for portable, on-device PLLM deployment.

% \vspace{-4mm}
\subsection{Internalized Memory Reading}
\label{sec:mem_reading}
While the dynamic memory writing produces a dense, redundancy-free, and temporally grounded external memory, off-the-shelf LLMs treat retrieved observations as disconnected \textit{``foreign texts''} instead of intrinsic personalization priors.  
To realize genuine personalization, we must \textit{internalize the user's context into the model's parameters} while utilizing episodic observations for precise factual grounding. This mirrors cognitive offloading \citep{risko2016cognitive, sweller1988cognitive}, akin to a student internalizing core  schemas while referencing a cheat sheet for specific formulas.

To achieve this without the prohibitive inference costs of full-context exposure or the train-inference gap of naive autoregressive training (\Cref{sec:prelim}), \model employs a self-learning pipeline that curates synthetic user data to compress extensive user experiences directly into LLM's parameters (\Cref{sec:self_learning}). We augment this with a teacher-student context distillation strategy~\citep[\Cref{sec:context_distillation}; ][]{snell2022learning, choi-etal-2023-fixed} to deeply instill user's context. While context distillation was originally designed to internalize detailed prompt instructions for reduced latency, to our knowledge,  we are the first to repurpose it to overcome the data scarcity and computational bottlenecks inherent in deep personalization~\citep{zhang-etal-2025-prime}.

\input{Tables/main_results}
\vspace{-2mm}
\subsubsection{Self-Learning on Synthetic Data}
\label{sec:self_learning}
Because user-specific labeled data is inherently scarce, existing models often rely on cross-user data~\citep{yan2025memory, wang2025mem}, raising significant privacy concerns. We bypass this by employing a self-learning paradigm~\citep{zhang2019your, pham2022revisiting, lee-etal-2025-self}, where \model is trained on synthetic question-answer (QA) pairs $\{(x,y)\}$, exclusively generated from the target user's context by itself.

Sepcfically, we prompt the LLM to generate 5W1H-based QA pairs for each conversational session, ensuring questions are meaningful and answers remain concise and specific. 
Next, we apply a data filtering pipeline to improve the synthetic data quality by removing: 1) trivial questions where the answer explicitly appears in the question; 2) non-appropriate QA pairs (\Cref{appx:non_appropriate_qa}); and 3) non-cycle-consistent QA pairs~\citep{alberti-etal-2019-synthetic} where the synthesized answer $y$ deviates from the teacher’s output $\hat{y}$.\footnote{We discard pairs if the semantic similarity is below 0.5.} 
This process produces a user-specific QA dataset $\mathcal{D}$ for subsequent context distillation (\Cref{sec:context_distillation}).
Notably, self-learning on our curated $\mathcal{D}$ \textbf{bridges the train–inference gap by} enabling $\mathcal{M}_{\theta_S}$ to internalize personalization signals through ``learn to answer'' instead of merely rote memorization of past conversations.

\vspace{-2mm}
\subsubsection{Context Distillation}
\label{sec:context_distillation}
During distillation, the \textit{teacher} $\mathcal{M}_{\theta_T}$ processes the full, uncompressed conversation history $C$ alongside the target question $x$, to generates a highly personalized response $\hat{y}$. The \textit{student} $\mathcal{M}_{\theta_S}$, \textbf{utilizing exactly the same architecture}, is strictly constrained to see the question $x$ only during training without any retrieved memory. Formally, keeping the teacher's parameters $\theta_T$ \textit{frozen}, we optimize the student's parameters $\theta_S$ to match the teacher's output distribution $\mathcal{M}_{\theta_T}(\cdot|[C; x])$ via a per-token KL divergence objective~\citep{cover1999elements} as follows:

\vspace{-4mm}

{\small
\begin{equation}
\begin{split}
\mathcal{L}_{\mathcal{D}, \theta_T}(\theta_S) & = \mathbb{E}_{(x,y)\sim \mathcal{D}} \Bigg[\mathbb{E}_{\hat{y}\sim \mathcal{M}_{\theta_T}(\cdot|x)} \sum_{n=1}^{|\hat{y}|} \\ 
&\quad \text{KL}\Big(\mathcal{M}_{\theta_T}(\cdot | [C; x], \hat{y}_{<n}) \;\Big\|\; \mathcal{M}_{\theta_S}(\cdot | x, \hat{y}_{<n})\Big)\Bigg]
\end{split}
\label{eq}
\end{equation}}
where $|\hat{y}|$ is the teacher's response length, $[\cdot;\cdot]$ is the textual concatenation operator, and $\hat{y}_{<n}$ is the tokens generated up to step $n$. Since computing the exact full-vocabulary per-token KL divergence is intractable, we approximate it on the \textit{top-$d$ tokens}, and we analyze the effect of $d$ in~\Cref{sec:further_analysis}. 

By aligning the student's probability space with the teacher's informed distribution, the student systematically learns to internalize the user experiences and preferences embodied in $\mathcal{D}$ into its own weights, even though \textit{it never explicitly receives the target user's exact conversation history during training}. This elegantly \textbf{resolves the quality-efficiency tradeoff} by decoupling the historical context length from the inference-time compute.

%% file: Tables/main_results.tex
\begin{table*}[t]
\centering
\vspace{-8mm}
\resizebox{0.95\linewidth}{!}{%
\begin{tabular}{lrrrrrrrrrrrr}  \toprule
\multirow{2}{*}{Method}                     & \multicolumn{3}{c}{Qwen3-4B}                                                      & \multicolumn{3}{c}{Qwen3-8B}                                                      & \multicolumn{3}{c}{Qwen3-14B}                                                     & \multicolumn{3}{c}{Qwen3-32B}                                                     \\  \cmidrule(lr){2-4}  \cmidrule(lr){5-7} \cmidrule(lr){8-10}  \cmidrule(lr){11-13}
                                            & {F1} & {BLEU-1} & {ROUGE-L} & {F1} & {BLEU-1} & {ROUGE-L} & {F1} & {BLEU-1} & {ROUGE-L} & {F1} & {BLEU-1} & {ROUGE-L} \\ \midrule
RAG (utterance)                             & 28.53                  & 23.20                      & 27.66                       & 28.10                  & 22.53                      & 27.30                       & 31.38                  & 26.02                      & 30.59                       & 28.64                  & 23.19                      & 27.83                       \\
RAG (observation)                           & 31.17                  & 25.50                      & 30.31                       & 31.77                  & 25.91                      & 30.63                       & 34.83                  & 28.79                      & 33.43                       & 34.00& 28.00& 32.56\\  \hdashline[5pt/4pt]
Vanilla Training                            & 5.76                   & 4.02                       & 5.53                        & 6.07                   & 4.32                       & 5.73                        & 6.59                   & 4.96                       & 6.24                        & 8.28                   & 6.22                       & 7.91                        \\
\model (no grounding)        & 12.26                  & 9.50                       & 11.84                       & 11.99                  & 9.33                       & 11.67                       & 14.71                  & 11.90                      & 14.41                       & 14.81                  & 11.27                      & 14.37                       \\  \hdashline[5pt/4pt]
\model (static accumulation) & \textbf{32.82}         & \textbf{27.21}             & \textbf{31.96}              & \textbf{34.50}         & \textbf{29.25}             & \textbf{33.40}              & 37.07                  & 31.25                      & 35.65                       & 32.75& 26.18& 31.46\\
\model (evolving memory)     & 31.38                  & 26.23                      & 30.40                       & 33.99                  & 28.92                      & 32.79                       & {\ul \textbf{37.63}}   & {\ul \textbf{32.16}}       & {\ul \textbf{36.16}}        & \textbf{34.25}& \textbf{28.12}& \textbf{32.85}\\ \bottomrule           
\end{tabular}
}
\vspace{-2mm}
\caption{Main results of \model and baselines on LoCoMo dataset (average of 3 runs), isolating module contributions. For each base model, the best results are in \textbf{bold}, while global optimal performances are {\ul underlined}. The results demonstrate information density of observations, usefulness of internalized memory reading, necessity for factual grounding, and efficacy of evolving memory. Complete results are shown in~\Cref{tbl:results_complete} including efficiency.}
\vspace{-6mm}
\label{tbl:results_main}
\end{table*}

%% file: 5_Experiment.tex
\input{Tables/SOTA_comparison}

\section{Experiments}
\label{sec:experiment}
\paragraph{Datasets and Tasks.} We conduct a holistic evaluation of PLLMs using the LoCoMo dataset~\citep{maharana-etal-2024-evaluating}. LoCoMo comprises long-term conversations spanning several months across 10 participant pairs. On average, each pair engages in $19.3$ sessions and $304.9$ turns. Following recent literature~\citep{chhikara2025mem0, yan2025memory}, we restrict our evaluation to the last eight participant pairs to ensure fair comparisons.
This evaluation set contains $1,294$ QA items.\footnote{While the initial set contains $1,307$ QA pairs, we filter out questions whose answers cannot be grounded in conversations.}
We further evaluate against SOTA models on LongMemEval~\citep{DBLP:conf/iclr/WuWYZCY25}, a benchmark of extremely long conversations between user and AI assistant, with an average of $47.7$ sessions and $493.4$ turns per user-assistant pair. This evaluation set contains $500$ QA items, with one unique test item per user profile.

Data creators confirm their datasets are free from personally identifiable info or offensive contents.

\paragraph{Setup.} We instantiate \model using recent, strong open-weight LLMs, \textsc{Qwen3}~\citep{yang2025qwen3} \textit{without thinking}. To ensure a comprehensive analysis, we experiment with models from 4B to 32B parameters.   Our primary results utilize $k=3$ retrieved memory entries, and the retrieval is conducted by \textsc{Qwen3-Embedding}-8B . During inference, we perform greedy decoding for reproducibility. Training details are listed in~\Cref{appx:train}.

We benchmark against \textsc{Qwen3}-8B~\citep{yang2025qwen3} specifically when compared with \textit{existing baselines}, 
since the SOTA method, Memory-R1~\citep{yan2025memory}, utilizes comparable-sized models like \textsc{Qwen2.5}-7B~\citep{DBLP:journals/corr/abs-2412-15115}. Notably, we report results for \model-PRO, which uses a more robust inference prompt and optimized hyper-parameters (detailed in~\Cref{appx:pro}).

\paragraph{Baselines and \model Variants.}  
We evaluate \model against two vanilla RAG baselines (retrieving raw utterances vs. extracted observations) and two training-based baselines (standard fine-tuning on raw conversations vs. a RAG-free variant of \model). Within the \model framework, we compare \textit{static accumulation} with \textit{evolving memory}, both built upon observations.

Further, we compare \model with several established baselines for long-horizon reasoning: LoCoMo~\citep{maharana-etal-2024-evaluating}, A-Mem~\citep{xu2025mem}, Mem0~\citep{chhikara2025mem0}, and Memory-R1~\citep[\textsc{Qwen2.5}-7B;][]{yan2025memory}.

\paragraph{Evaluation Metrics.} We employ six quality metrics to evaluate model performance: token-level Precision (P), Recall (R), and F1-score, alongside BLEU-1, ROUGE-L, and LLM-as-a-Judge (J). Except for J, these metrics quantify the lexical overlap between the generated and ground-truth answers. Following~\citet{yan2025memory}, we utilize GPT-5-nano~\citep{singh2025openai} for the J metric to assess semantic correctness, relevance, completeness, and contextual appropriateness. \textit{J  is  only used for SOTA comparison to minimize costs}.
We also report inference efficiency measured by input token length.

%% file: Tables/SOTA_comparison.tex
\begin{table*}[t]
\centering
\vspace{-8mm}
\resizebox{0.95\linewidth}{!}{%
\begin{tabular}{lrrrrrrrrrrrrrrr}    \toprule
  \multirow{2}{*}{Method}    & \multicolumn{3}{c}{Single Hop}                                              & \multicolumn{3}{c}{Multi-Hop}                                               & \multicolumn{3}{c}{Open Domain}                                             & \multicolumn{3}{c}{Temporal}                                                & \multicolumn{3}{c}{Overall}                                                 \\  \cmidrule(lr){2-4} \cmidrule(lr){5-7}  \cmidrule(lr){8-10} \cmidrule(lr){11-13}   \cmidrule(lr){14-16}
                                                 & {F1} & {BLEU-1} & {J} & {F1} & {BLEU-1} & {J} & {F1} & {BLEU-1} & {J} & {F1} & {BLEU-1} & {J} & {F1} & {BLEU-1} & {J} \\   \midrule
LoCoMo (\citeauthor{maharana-etal-2024-evaluating})                                           & 9.57                   & 7.00                       & 15.06                 & 11.84                  & 10.02                      & 19.28                 & 8.67                   & 6.52                       & 12.79                 & 8.35                   & 8.74                       & 5.43                  & 8.97                   & 7.27                       & 12.17                 \\
A-Mem (\citeauthor{xu2025mem})                                            & 18.96                  & 12.86                      & 40.78                 & 14.73                  & 12.66                      & 31.32                 & 30.58                  & 26.14                      & 46.90                 & 23.67                  & 20.67                      & 28.68                 & 26.08                  & 21.78                      & 40.78                 \\
Mem0 (\citeauthor{chhikara2025mem0})                                             & 24.96                  & 18.05                      & 61.92                 & 20.31                  & 15.82                      & 48.19                 & 32.74                  & 25.27                      & 65.20                 & 33.16                  & 26.28                      & 38.76                 & 30.61                  & 23.55                      & 53.30                 \\
Memory-R1-PPO (\citeauthor{yan2025memory})                                     & 34.22                  & 23.61                      & 57.74                 & 32.87                  & 29.48                      & 53.01                 & 44.78                  & 38.72                      & 66.99                 & 42.88                  & 30.30                      & 42.25                 & 41.72                  & 33.70                      & 59.53                 \\
Memory-R1-GRPO (\citeauthor{yan2025memory})                                   & 33.64                  & 26.06                      & 62.34                 & 23.55                  & 20.71                      & 40.96                 & \textbf{46.86}         & \textbf{40.92}             & \textbf{67.81}& \textbf{47.75}         & 38.49                      & \textbf{49.61}& 43.14                  & 36.44                      & 61.51                 \\ \hdashline[5pt/4pt]
\model-PRO (static acc.)   & \textbf{50.87}         & \textbf{45.61}             & \textbf{84.92}& 37.39                  & 30.94                      & 80.61& 26.16                  & 23.43                      & 61.50& 42.14                  & 37.49                      & 33.69& \textbf{45.24}         & \textbf{40.02}             & \textbf{72.59}\\
\model-PRO (evolving mem.) & 48.26                  & 43.21                      & 81.68& \textbf{38.39}         & \textbf{32.71}             & \textbf{82.20}& 27.81                  & 24.85                      & 62.72& 45.82                  & \textbf{42.07}             & 42.19& 44.65                  & 39.87                      & 72.46\\ \bottomrule  
\end{tabular}
}
\vspace{-2mm}
\caption{Comparison of \model-PRO (average of 3 runs) with prior SOTA baselines. Results of non-\model baselines are taken from~\citet{yan2025memory}. Best numbers are in \textbf{bold}. Overall, \model-PRO consistently outperforms all baselines, including the highly competitive Memory-R1. The sole exception is the open-domain sub-task, where our performance is lower as the current pipeline is primarily optimized for factoid QA reasoning.}
\label{tbl:results_SOTA}
\vspace{-6mm}
\end{table*}

%% file: 6_Results.tex
\section{Results}

\subsection{Main Results and Systematic Evaluation}
\label{sec:main_results}

We systematically evaluate \model\ across multiple scales to isolate module contributions. Major results are included in~\Cref{tbl:results_main}, and complete results (including all quality metrics and efficiency) are in \Cref{tbl:results_complete}. Below are our major findings:

\begin{itemize} [leftmargin=1em,itemsep=0em,topsep=1pt,parsep=1pt,partopsep=1pt]
    \item \textbf{Information Density of Observations:} \texttt{RAG (observation)} consistently outperforms \texttt{RAG (utterance)}. This validates that compressing raw, noisy utterances into core factual observations creates higher-density retrieval targets, significantly enhancing response quality while reducing token budgets.

    \item \textbf{Efficacy of Internalized Memory Reading:} Our \textit{self-learning} pipeline effectively bridges the train-inference gap observed in \texttt{Vanilla Training} (\Cref{sec:prelim}). Further, KL-divergence-based \textit{context distillation} provides richer supervision to capture nuances than sparse cross-entropy, enabling more genuine personalization.  

    \item \textbf{Necessity of Explicit Grounding:} Removing access to external episodic memory causes severe performance collapse. The performance decline in both \texttt{Vanilla Training} and \model \texttt{(no grounding)} confirms that  parameterization alone to store user context is insufficient---explicit grounding is imperative for \textit{high-fidelity} personalization.
    
    \item \textbf{Impact of Evolving Memory:} Employing a memory manager to autonomously govern evolution yields peak quality scores (e.g., $37.63$ F1) while minimizing input lengths. However, this manager requires sufficient model capacity, realizing its full potential \textit{only beyond 8B params}.

\end{itemize}

\paragraph{Scaling Saturation at 32B.}
While \model\ significantly outperforms baselines at 4B-14B scales, it saturates at 32B, surpassing \texttt{RAG (observation)} by merely $0.7\%$. We hypothesize that larger models already possess sufficient intrinsic reasoning and in-context learning capabilities to robustly comprehend retrieved observations, often perceived as \textit{``foreign texts''} to smaller LLMs.

\begin{figure*}[t]
    \centering
    \vspace{-8mm}
    \includegraphics[width=0.9\textwidth]{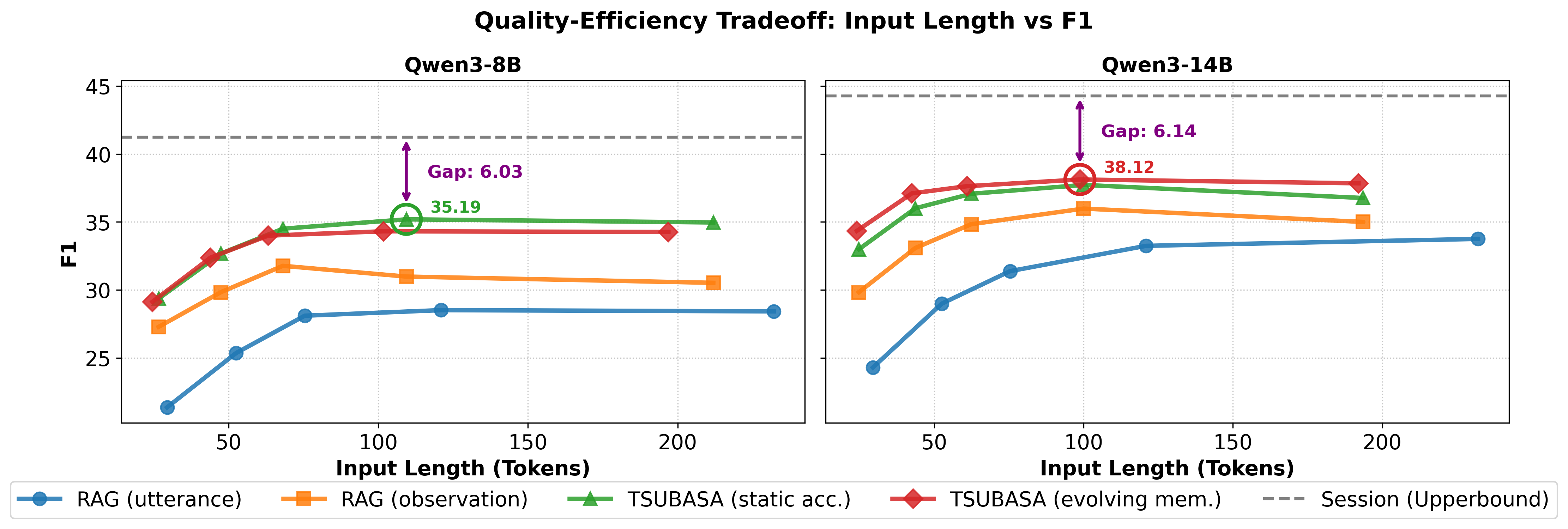}
    \vspace{-4mm}
    \caption{Quality-Efficiency tradeoff between input length and F1 metric on baselines and \model. Best-performing configuration is highlighted in circle with its gap to the establish ceiling (\texttt{Session}) indicated.
    Importantly, our \model achieves \textit{Pareto improvement}:  attaining higher peak performance while utilizing only a fraction of the context budget.
    Tradeoffs w.r.t. ROUGE-L and BLEU-1 are presented in~\Cref{fig:analysis1_r1} and~\Cref{fig:analysis1_b1}.}
    \label{fig:analysis1_f1}
    \vspace{-4mm}
\end{figure*}

\begin{figure}[t]
    \centering
    \includegraphics[width=0.5\textwidth]{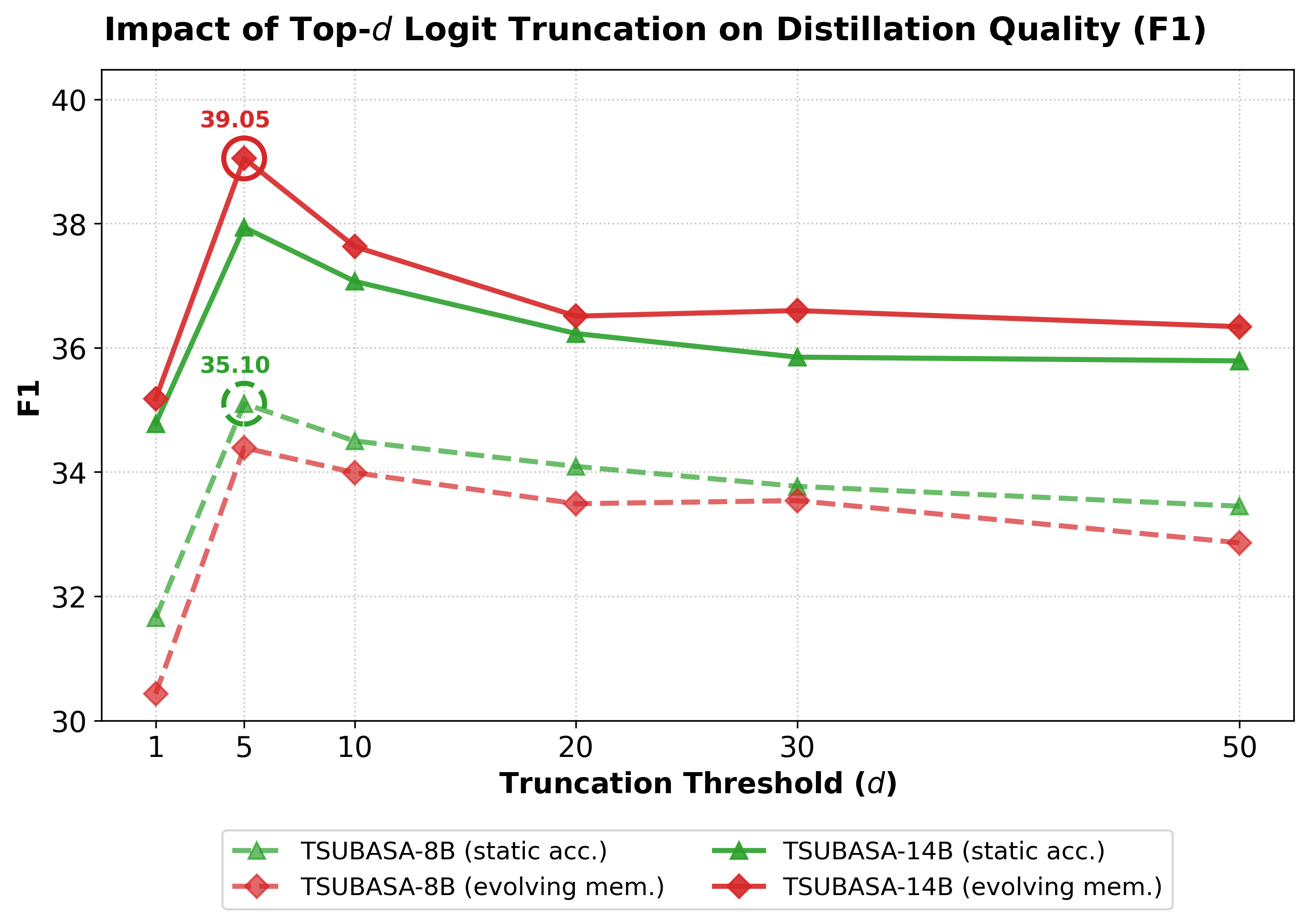}
    \vspace{-4mm}
    \caption{Impact of top-$d$ truncation for approximating KL divergence on distillation quality (F1) with \model variants.
    Best-performing config is highlighted in circle for each model size.
    Importantly, we find $d=1$ inadequate, while increasing $d$ beyond $5$ leads to performance degradation due to noises in long-tail distribution, establishing $d=5$ at the sweet spot.
See~\Cref{fig:analysis2_r1} and~\Cref{fig:analysis2_b1} for impacts on ROUGE-L and BLEU-1.}
    \label{fig:analysis2_f1}
    \vspace{-4mm}
\end{figure}

\subsection{Comparison with SOTA Baselines}
\label{sec:results_SOTA}

As shown in~\Cref{tbl:results_SOTA}, \model-PRO significantly outperforms existing baselines on LoCoMo benchmark, particularly in explicit \textit{fact-retrieval} tasks like single/multi-hop reasoning. However, we observe a performance degradation in the \textit{Open Domain} task (e.g., connecting a preference for classical music to a specific composer \textit{not mentioned} in the context).
This bottleneck is a direct consequence of our synthetic data generation pipeline, which is optimized for \textit{factoid QA reasoning} rather than \textit{abstract reasoning}. We acknowledge this limitation and advocate for future synthetic data generation pipeline to include more complex commonsense reasoning QA pairs to further bridge the gap.

Notably, \model reaches the best overall F1 of $45.24$ and J of $72.59$, not only critically surpassing the competitive baseline Memory-R1 but also outperforming industry standards like Mem0 by $35$-$50\%$. Importantly, unlike Memory-R1 which suffer from a serious privacy flaw by training on  other user's data, \model operates strictly within the target user's portfolio. This ensures robust performance without data privacy concerns.

\Cref{tbl:results_longmemeval} presents the LongMemEval results. \model-PRO attains performance comparable to Memory-R1 while using only 3 retrieved memory pieces compared to Memory-R1's 60. Meanwhile, it also surpasses the popular industry system mem0 by an average of over $20\%$ across three quality metrics, further demonstrating the strong competitiveness of \model.

\subsection{Further Analyses}
\label{sec:further_analysis}
\paragraph{Pareto improvement: Breaking the Quality-Efficiency Tradeoff.} 

Traditional retrieval methods often force a compromise between response quality and latency costs of long contexts. To evaluate how \model addresses this, we analyzed response quality---F1 (\Cref{fig:analysis1_f1}), BLEU-1 (\Cref{fig:analysis1_b1}), and ROUGE-L (\Cref{fig:analysis1_r1})---against input lengths for $k\in\{1,2,3,5,10\}$ retrieved memory pieces. 
As shown in \Cref{fig:analysis1_f1}, \model breaks this tradeoff and achieves a \textit{Pareto improvement}, reaching higher peak performance while utilizing only a fraction of the context budget. For example, with \textsc{Qwen3}-14B, \texttt{\model (evolving)} at $k=2$ achieves an F1 score of $37.10$ with only $42.5$ tokens. This easily surpasses the $k=10$ configuration of \texttt{RAG (utterance)}, which requires $232.08$ tokens to reach a $33.75$ F1 score, echoing the importance of \textit{information density} in retrieval targets as discussed in~\Cref{sec:main_results}.

While reasoning with the full session history (\texttt{Session}) represents a performance ceiling, it incurs a prohibitive cost of $2,158$ tokens. \model effectively closes this gap: by $k=5$ ($\sim100$ tokens), it captures the majority of relevant context,  approaching the ceiling with $20\times$ fewer tokens.

\paragraph{Sweet-Spot of Top-$d$ Truncation for Context Distillation}

Computing exact KL divergence over the entire vocabulary\footnote{For \texttt{Qwen3}, the vocab size exceeds $150k$.} for our context distillation objective is intractable. We approximate this by restricting the calculation to the top-$d$ tokens, sorted by probability mass. Evaluating truncation thresholds $d \in \{1, 5, 10, 20, 30, 50\}$, we observe a non-monotonic relationship with response quality (\Cref{fig:analysis2_f1}). This establishes $d=5$ as the sweet spot and reveals two insights: 

\begin{itemize} [leftmargin=1em,itemsep=0em,topsep=1pt,parsep=1pt,partopsep=1pt]
    \item \textbf{Inadequacy of Hard Labels ($d=1$):} Truncating at $d=1$ yields the worst performance, as KL divergence collapses into cross-entropy over the argmax token. This highlights the necessity of soft labels, as the probability distribution among top candidate tokens contains rich semantic nuances that better guide the student's learning.
    
    \item \textbf{Long-Tail Noise Corrupts Learning ($d \ge 10$):} Counter-intuitively, performance steadily declines as $d > 5$. While a wider distribution seemingly captures more knowledge, the vocabulary's long tail is usually populated with noisy or irrelevant tokens~\citep{li-etal-2025-bild}. Thus, aggressive truncation at $d=5$ acts as a vital regularizer, filtering out  teachers' noise while preserving essential dark knowledge.
\end{itemize}

%% file: 7_Conclusion.tex
\section{Conclusion}

In this work, we present \model to tackle the long-horizon personalization task. The framework of \model is motivated by limitations of current memory-augmented PLLMs: low-information-density retrieval targets in memory, train-inference gap, and quality-efficiency tradeoff. Specifically, \model features two \textit{wings}: \textit{dynamic memory writing} via structured algorithmic memory evolution on the basis of factual, information-dense observations, and \textit{internalized memory reading} through  self-learning pipeline with context distillation objective. We conduct extensive experiments to validate the design choice and quantify the contribution by each module. Benchmarking shows that \model surpasses significantly prior SOTA models in both quality and efficiency on LoCoMo.

%% file: Limitation.tex
\section*{Limitations}

\paragraph{Evaluation benchmarks.} In this work, we have included two evaluation benchmarks, aiming to cover a diverse array of long-horizon personalization tasks. Yet, these two benchmarks cannot comprehensively represent the entire spectrum. For future research, we plan to extend \model to more applications, and examine its true generalizability in the wild.  

\paragraph{Model Scales.} In this study, we evaluate a diverse set of models ranging from 4$B$ to 32$B$ parameters. Owing to budget constraints, we do not extend our experiments to larger-scale models.
Nonetheless, we expect our findings to hold on such models, and we have also discussed potential saturation effect post-$32B$ (\Cref{sec:main_results}). We leave the direct application of \model to larger models in the future work.

\paragraph{GPU resources.} The base LLMs used in this work are of 4 to 32 billions parameters. 
It is thus more time-consuming than traditionally small models like BERT \citep{devlin-etal-2019-bert} at both training and inference time, which in turn results in a higher carbon footprint.
Specifically, for training, we run each base model of size 4$B$ to 14$B$ on 2 NVIDIA L40, and the $32B$ model on 2 NVIDIA RTX PRO 6000. for inference, we run \model on 1 single NVIDIA L40. All come with significant CPU and memory resources. 
The training time is generally 1-2 minutes per epoch. The inference time for each LLM on each benchmark ranges from several minutes to several hours, depending on the configurations.

%% file: Appendix.tex
\section{Non-appropriate QA pairs}
\label{appx:non_appropriate_qa}
Taking LoCoMo for instance, during our data filtering process within the self-learning pipeline, we consider the following three categories as non-appropriate QA pairs.
\begin{itemize} [leftmargin=2em,itemsep=0em,topsep=1pt,parsep=1pt,partopsep=1pt]
\item Uninformative temporal questions (e.g., asking for the timestamp of a chat message rather than when an actual activity occurred)
\item Unanswerable questions (e.g., the synthetic reference answer contains phrases like ``i do not know'', ``not mentioned'',  ``unspecified'', etc.)
\item Overly long answers (i.e., here, we heuristically drop QA pairs whose answer contains more than 30 tokens.)
\end{itemize}

\section{Training Details of Internalized Memory Reading}
\label{appx:train}
When performing parameter updating (\Cref{sec:mem_reading}) using LoRA technique~\citep{hu2022lora}, we manually set 5 epochs with warm-up ratio of $0.05$, batch size at $4$, gradient accumulation steps at $2$, learning rate at $5\times10^{-4}$ and rank $r=16$, and we systematically study the impact of $d$ (default: 10) for approximating KL divergence in~\Cref{sec:further_analysis}, while leaving other hyper-parameters as default.

\section{Comparison between \model and \model-PRO}
\label{appx:pro}

Exactly the same framework and underlying models are used for both versions. The training details are identical for both, as described in~\Cref{sec:experiment}, except for $d$ in KL divergence estimation. $d=10$ and $d=20$ are used by \model and \model-PRO, respectively. The major differences reside in inference-time: 1) Compared with the system prompt designed for \model, \model-PRO additionally emphasizes on \textit{concise generation}, as inspired by~\citet{yan2025memory}; 2) the number of retrieved items ($k$) are different: \model sets $k=3$ by default, while \model-PRO sets $k=10$.

\input{Tables/SOTA_longmemeval}

\begin{figure*}[t]
    \centering
    \includegraphics[width=0.8\textwidth]{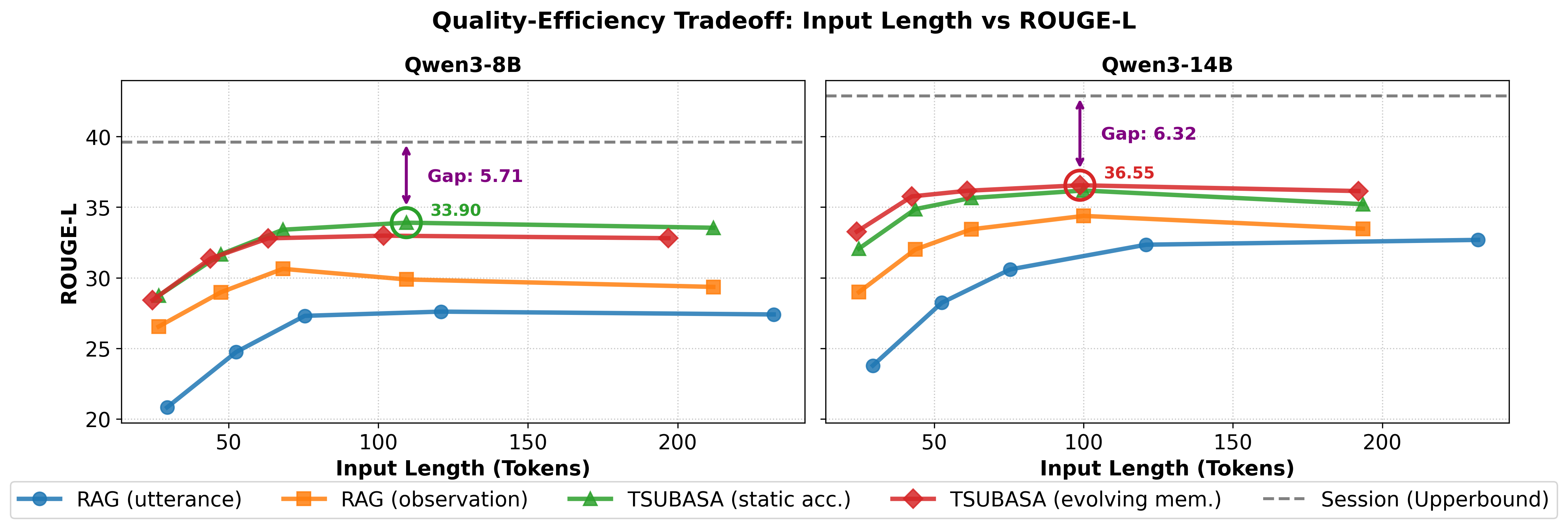}
    \caption{Quality-Efficiency tradeoff between input length and ROUGE-L metric on baselines and \model. Best-peroforming configuration is highlighted in circle with its gap to the establish ceiling (\texttt{Session}) indicated.
    Importantly, our \model achieves \textit{Pareto improvement}:  attaining higher peak performance while utilizing only a fraction of the context budget.}
    \label{fig:analysis1_r1}
\end{figure*}

\begin{figure*}[t]
    \centering
    \includegraphics[width=0.8\textwidth]{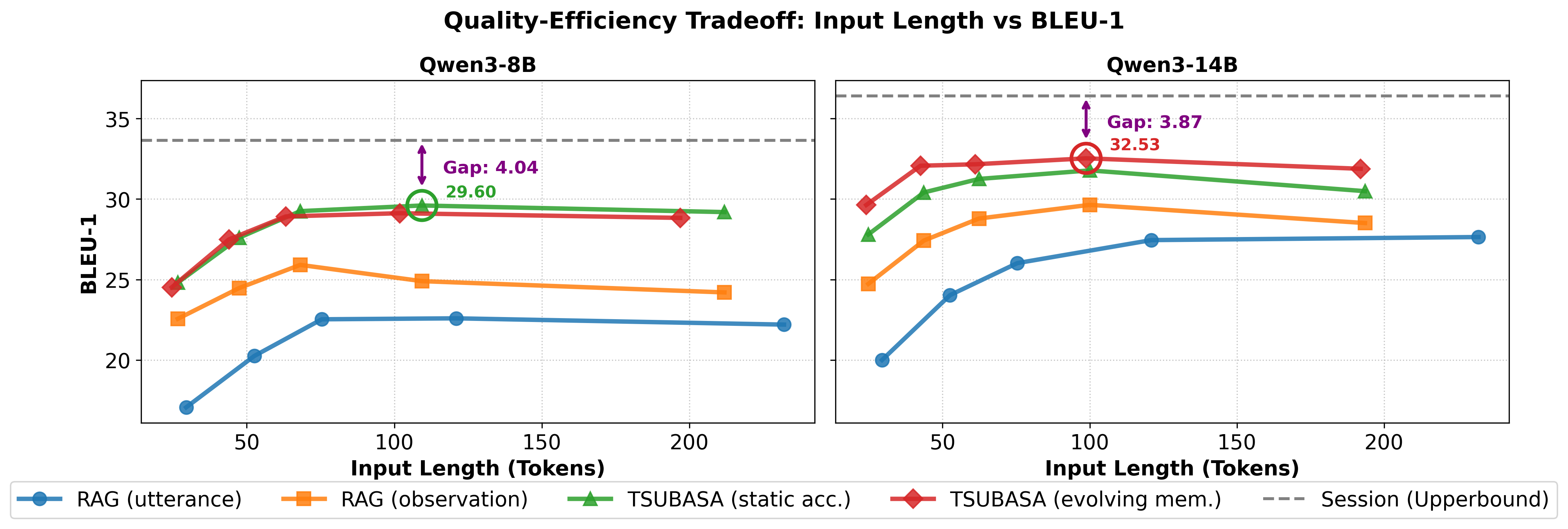}
    \vspace{-4mm}
    \caption{Quality-Efficiency tradeoff between input length and BLEU-1 metric on baselines and \model. Best-peroforming configuration is highlighted in circle with its gap to the establish ceiling (\texttt{Session}) indicated.
    Importantly, our \model achieves \textit{Pareto improvement}:  attaining higher peak performance while utilizing only a fraction of the context budget.}
    \label{fig:analysis1_b1}
\end{figure*}

\begin{figure}[t]
    \centering
    \includegraphics[width=0.45\textwidth]{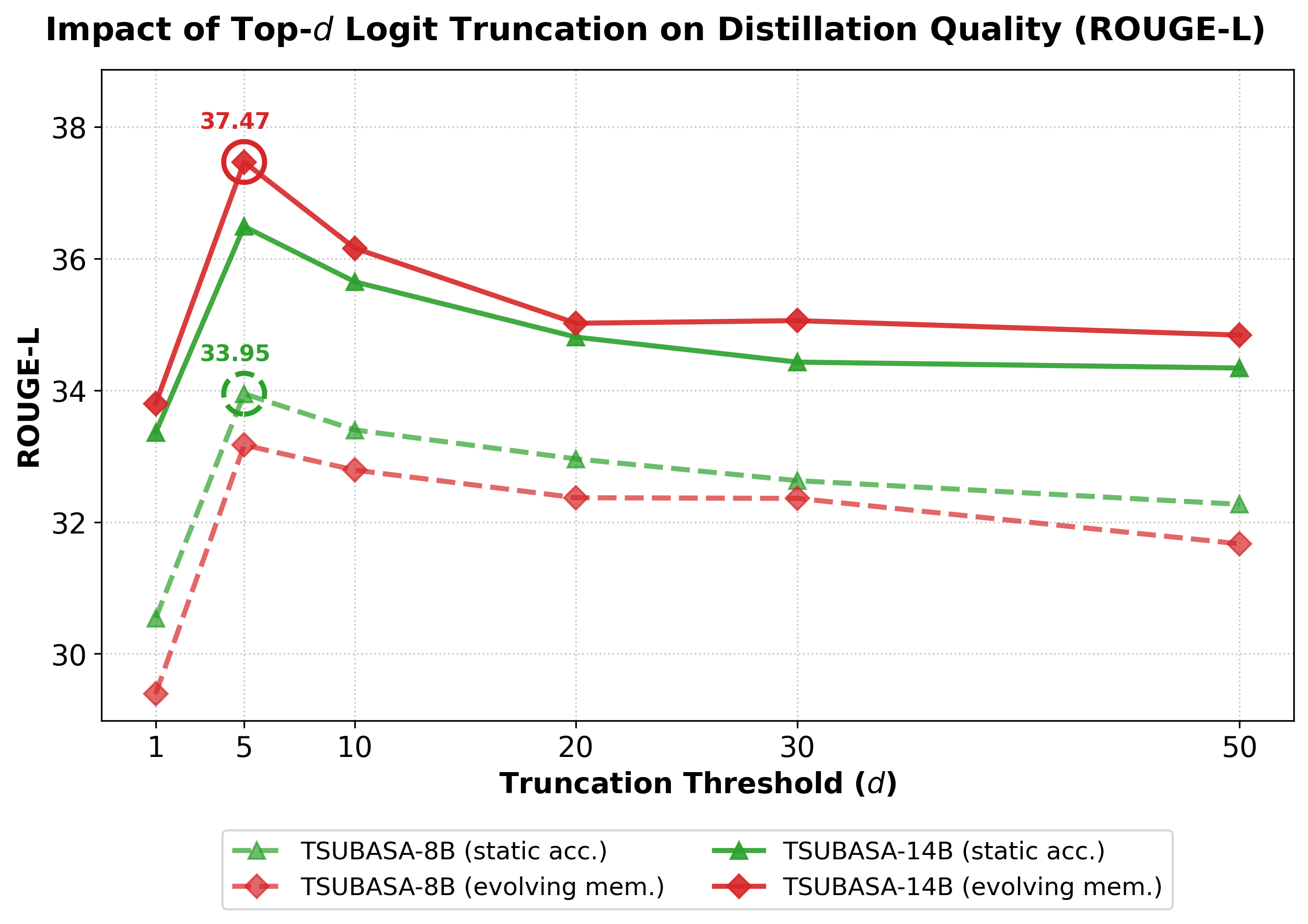}
    \caption{Impact of top-$d$ truncation for approximating KL divergence on distillation quality (ROUGE-L) with \model variants.
    Best-performing configuration is highlighted in circle for each model size.
    Importantly, we find $d=1$ inadequate, while increasing $d$ beyond $5$ leads to performance degradation due to noises in long-tail distribution, establishing $d=5$ at the sweet spot.}
    \label{fig:analysis2_r1}
\end{figure}

\begin{figure}[t]
    \centering
    \includegraphics[width=0.45\textwidth]{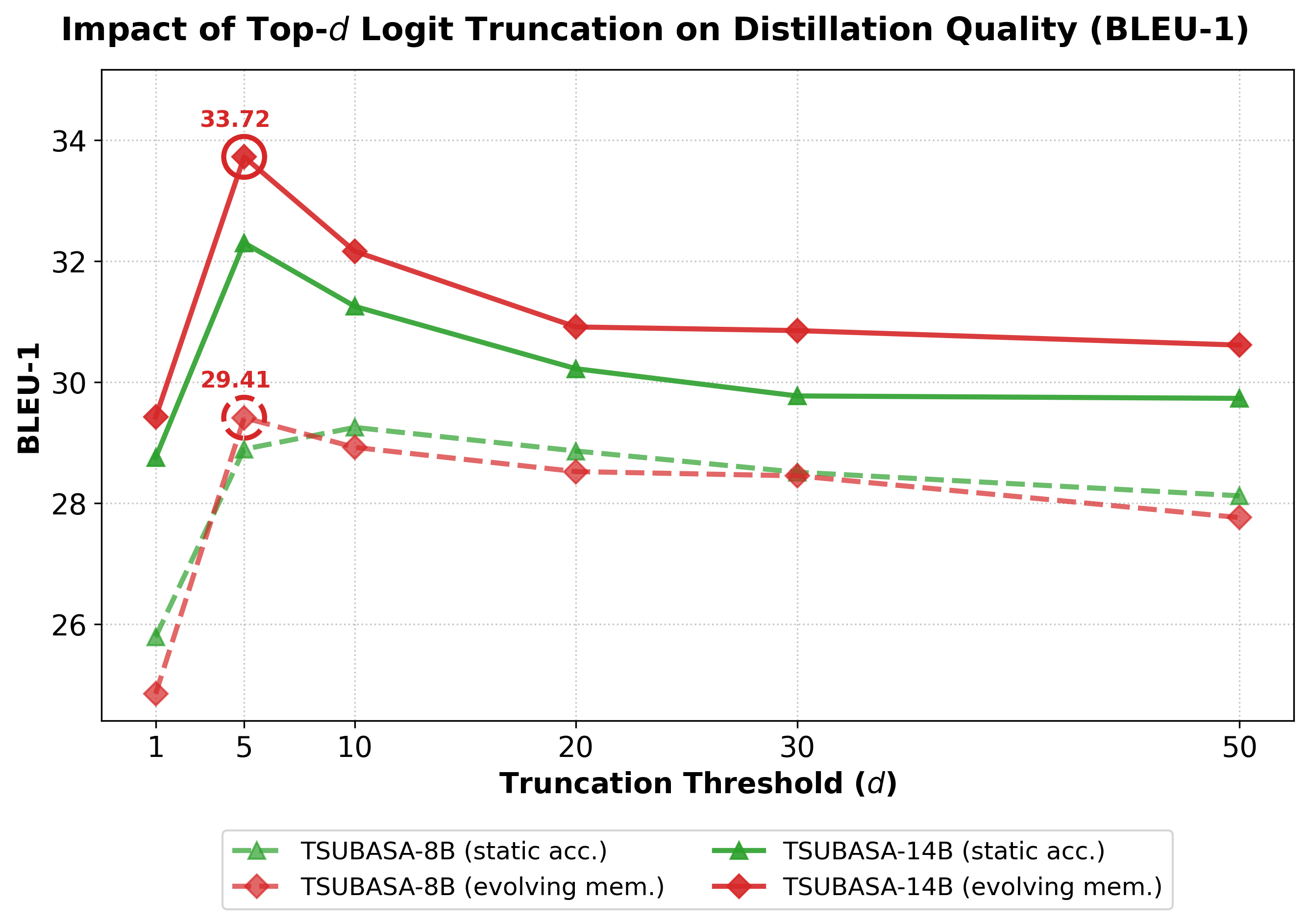}
    \caption{Impact of top-$d$ truncation for approximating KL divergence on distillation quality (BLEU-1) with \model variants.
    Best-performing configuration is highlighted in circle for each model size.
    Importantly, we find $d=1$ inadequate, while increasing $d$ beyond $5$ leads to performance degradation due to noises in long-tail distribution, establishing $d=5$ at the sweet spot.}
    \label{fig:analysis2_b1}
\end{figure}

\input{Tables/complete_results}

\begin{figure*}[t]    
  \centering
  \small
\begin{tcolorbox}[
    enhanced,
    colback=white,
    colframe=black,
    coltitle=white,
    fonttitle=\bfseries,
    title=Observation Extraction Template,
    sharp corners,
    boxrule=1pt,
    titlerule=0pt,
    toptitle=1.5mm,
    bottomtitle=1.5mm,
    left=4mm,
    right=4mm,
    top=4mm,
    bottom=4mm
]

Analyze the conversation between \{speaker\_a\} and \{speaker\_b\}, and extract the major OBSERVATIONS for \{speaker\_target\}.

\vspace{1em}
\noindent \textbf{Core Requirement:} The OBSERVATIONS are defined as verifiable facts about \{speaker\_target\}'s life, background, and habits. Avoid abstract ``meta-observations'' about the communication dynamics such as `\{speaker\_target\} is supportive', `\{speaker\_target\} appreciates' etc; Instead, focus on concrete facts such as past events, recurring habits, and stated preferences of \{speaker\_target\}.

\vspace{1em}
\noindent \textbf{Organic 5W1H Integration:} Each OBSERVATION should be a standalone sentence that provides as much salient context as possible while adhering to the 5W1H principle. While not every OBSERVATION requires all six points, strive to include:
\begin{itemize}
    \item[-] \textbf{Who/What:} The specific entity, role, or action.
    \item[-] \textbf{Where:} The location or setting associated with the fact.
    \item[-] \textbf{When:} The timeframe, frequency, or specific date.
    \item[-] \textbf{Why/How:} The stated reason or method behind an action or preference.
\end{itemize}

\vspace{0.5em}
\noindent \textbf{Temporal Awareness:} The timestamp provided (\{time\}) reflects only when the conversation took place. Do not assume events discussed happened on this date unless explicitly stated (e.g., ``I did this yesterday''). Distinguish between the date of the report and the date of the event.

\vspace{1.5em}
\noindent ------------------------- \\
\{conversation\} \\
-------------------------

\vspace{1.5em}
\hrule
\vspace{1.5em}

Return a JSON object of the form \{\{ ``OBSERVATIONS'': [``<string>'', ``<string>'', ...] \}\}. 
The value should be a list of strings, each string being a major OBSERVATION about \{speaker\_target\}.
Do not include any explanations, commentary, or text outside the JSON object. 
Important: Escape all double-quote characters within string output with backslash.
End your response with ``\#END''.

\end{tcolorbox}
    \caption{Prompt used for extracting salient observations from a conversation session.}
    \label{fig:prompt_extraction}
\end{figure*}

\begin{figure*}[t]    
  \centering
  \small
\begin{tcolorbox}[
    enhanced,
    colback=white,
    colframe=black,
    coltitle=white,
    fonttitle=\bfseries,
    title=Memory Evolution Prompt,
    sharp corners,
    boxrule=1pt,
    titlerule=0pt,
    toptitle=1.5mm,
    bottomtitle=1.5mm,
    left=4mm,
    right=4mm,
    top=4mm,
    bottom=4mm,
]

You are a Memory Manager. Your task is to evolve a person's ``Memory Store'' based on a list of ``New Observations'' extracted from a recent conversation session. Your goal is to maintain a ``Memory Store'' that is optimized for Vector Database Retrieval.

\vspace{1em}
\noindent \textbf{RETRIEVAL PRINCIPLE:}\\
A retriever must be able to understand a ``refined\_observation'' without seeing any other context.

\vspace{1em}
\noindent \textbf{ACTIONS:}
\begin{enumerate}
    \item \textbf{``ADD''}: Brand new information.
    \item \textbf{``UPDATE''}: New info adds specific detail or a more recent status to an existing index.
    \item \textbf{``RECONCILE''}: New info contradicts or changes an existing state. Combine them into a ``Temporal Narrative'' (e.g., ``Subject previously X, but as of [Date] now Y'').
    \item \textbf{``IGNORE''}: The info is already perfectly captured or is trivial chatter.
\end{enumerate}

\vspace{1em}
\noindent \textbf{CURRENT MEMORY STORE for \{speaker\}:}\\
\{current\_memory\}

\vspace{1em}
\noindent \textbf{NEW OBSERVATIONS LIST for \{speaker\} [session date : \{timestamp\} (context: today)]:} \\
\{new\_obs\_list\}

\vspace{1em}
\noindent \textbf{RULES FOR ``refined\_observation'':}
\begin{itemize}
    \item \textbf{Sequential Processing:} You must process every observation in the ``NEW OBSERVATIONS LIST'' in order.
    \item \textbf{Temporal Grounding:} You MUST include both the absolute date and the relative timeframe (e.g., ``last week'', ``yesterday''), if either is mentioned in the observation. Perform calculations relative to the session date. This is crucial for retrieval relevance.
    \begin{itemize}
        \item[-] \textit{Example}: If the session date is 25 February, 2026, and \{speaker\} says ``I bought a car two days ago,'' the refined observation must include ``On 23 February, 2026 (context: two days ago), \{speaker\} bought a car.''
    \end{itemize}
    \item \textbf{Stand-alone Density:} Each refined observation must be a complete, keyword-rich sentence, that defines the Subject, Action, and Context.
    \item \textbf{Index Integrity:} If ``UPDATE'' or ``RECONCILE'', the ``index'' must be the exact integer from the current store.
\end{itemize}

\vspace{1.5em}
\hrule
\vspace{1.5em}

Return your response in STRICT JSON format. The output must be a list of objects:
\begin{verbatim}
[
  {
    "original_obs": "The raw observation from the NEW OBSERVATIONS LIST",
    "action": "ADD" | "UPDATE" | "RECONCILE" | "IGNORE",
    "index": index_to_change_or_null,
    "refined_observation": "The updated or new string to store. It must be self-contained 
                           and standalone."
  },
  ...
]
\end{verbatim}
\#END

\end{tcolorbox}
    \caption{Prompt used by our memory manager to autonomously evolve the episodic memory.}
    \label{fig:prompt_evolution}
\end{figure*}

\begin{figure*}[htpb]
\hspace{-7mm}
% \centering
\small
\begin{minipage}{1.25\textwidth}
\scalebox{0.85}{
\begin{tcolorbox}[
    colback=white,
    colframe=black,
    colbacktitle=black,
    coltitle=white,
    fonttitle=\bfseries,
    title=Dynamic Memory Writing Process: Observation Extraction and Memory Evolution,
    arc=0mm, % Sharp corners
    boxrule=0.75pt,
    left=3mm, right=3mm, top=3mm, bottom=3mm
]

\textbf{RAW UTTERANCES (Session 6)} \\
\textit{Timestamp: 2:33 pm on 5 February, 2023}
\vspace{1mm}

\dots

\textbf{John:} Yeah, Maria, we learn a lot from our own struggles. I just started helping out with a food drive for folks who lost their jobs. Here's a picture of me at the food bank.

\textbf{Maria:} Wow, John, that's incredible! What inspired you to get involved with something like this?

\textbf{John:} Seeing the effect \red{unemployment has on our neighbors} made me decide to act. I wanted to help out in these tough times by \red{doing a community} \red{food drive}. We can all make a difference!

\textbf{Maria:} That's really great of you. What sparked your decision to start this initiative?

\textbf{John:} Thanks, Maria. Unemployment in our community was the reason behind it.

\textbf{Maria:} You did awesome! How's the response been to that?

\textbf{John:} Thanks, Maria! We've been \green{overwhelmed by the response and the volunteers}. Here's a photo of them at a recent event.

\dots

\vspace{2mm}
\hrule
\vspace{2mm}

\textbf{EXTRACTED OBSERVATIONS (Session 6)} \\
\textit{Timestamp: 2:33 pm on 5 February, 2023}
\vspace{1mm}
\begin{itemize}
    \setlength{\itemsep}{0pt}
    \item \dots
    \item John is involved in political activities as he mentioned running for office to make an impact.
    \item \red{John initiated a community food drive due to the impact of unemployment on his neighbors.}
    \item \green{John has been overwhelmed by the positive response and volunteer participation in his food drive initiative.}
    \item \dots
\end{itemize}

\noindent\tikz \draw [dashed] (0,0) -- (\linewidth,0);

\textbf{EXTRACTED OBSERVATIONS (Session 28)} \\
\textit{Timestamp: 5:19 pm on 5 August, 2023}
\vspace{1mm}
\begin{itemize}
    \setlength{\itemsep}{0pt}
    \item \dots
    \item John lost his job at a mechanical engineering company on 5 August, 2023, due to the company tanking, which he did not foresee, and it has been a rough period for him.
    \item \blue{John has not been able to volunteer much lately, but he still cares about volunteering and values its impact on the community.}
    \item \dots
\end{itemize}

\vspace{2mm}
\hrule
\vspace{2mm}

\textbf{EVOLVING MEMORY STORE} \textit{(Action: RECONCILE)}
\vspace{1mm}

\begin{itemize}
    \setlength{\itemsep}{0pt}
    \item \dots
    \item \purple{$[22]$} \red{On 5 February, 2023, John initiated a community food drive due to the impact of unemployment on his neighbors.} \blue{On 5 August, 2023,} \blue{John has not been able to volunteer much lately, but he still cares about volunteering and values its impact on the community.}
    \item \dots
\end{itemize}

\end{tcolorbox}
}
\end{minipage}
\caption{A sample example illustrating the \model memory writing process. Raw utterances are first compressed into dense observations, which are subsequently merged by the memory manager (RECONCILE action in this particular example) to form a coherent, temporally grounded narrative. \textbf{The colored text indicates the flow.} For example, there are two observations that can be extracted from the truncated raw conversations, one in \red{red} and another in \green{green}. The \purple{sampled entry (index: 22)} from our evolving memory is derived from observations in \red{session 6} and \blue{session 28}. Note, this \purple{evolving memory piece} is temporally grounded with  absolute time markers.}
\label{fig:memory_example}
\end{figure*}

%% file: Tables/SOTA_longmemeval.tex
\begin{table}[t]
\centering
\resizebox{0.95\linewidth}{!}{%
\begin{tabular}{lrrr} \toprule
Method                                           & {F1} & {BLEU-1} & {J} \\ \midrule
LoCoMo (\citeauthor{maharana-etal-2024-evaluating})                                         & 18.27                  & 14.57                      & 22.20                 \\
A-Mem (\citeauthor{xu2025mem})                                           & 41.55                  & 36.58                      & 54.80                 \\
Mem0 (\citeauthor{chhikara2025mem0})                                           & 38.44                  & 34.53                      & 46.80                 \\
Memory-R1-PPO (\citeauthor{yan2025memory})                                   & 40.30                  & 35.50                      & 47.40                 \\
Memory-R1-GRPO (\citeauthor{yan2025memory})                                  & \textbf{46.70}         & 41.10                      & \textbf{57.80}        \\ \hdashline[5pt/4pt]
\model-PRO (static acc.)    & 45.75                  & \textbf{42.10}             & 57.40                 \\
\model-PRO (evolving mem.) & 43.43                  & 39.85                      & 53.00      \\  \bottomrule          
\end{tabular}
}
\vspace{-2mm}
\caption{Comparison of \model-PRO (single run) with prior SOTA baselines on LongMemEval~\citep{DBLP:conf/iclr/WuWYZCY25}. Results of non-\model baselines are taken from~\citet{yan2025memory}. Best numbers are in \textbf{bold}. Overall, \model-PRO consistently outperforms all baselines, including the highly competitive Memory-R1.}
\label{tbl:results_longmemeval}
\end{table}

%% file: Tables/complete_results.tex
\begin{table*}[t]
\centering
\resizebox{0.95\linewidth}{!}{%
\begin{tabular}{llrrrrrr} \toprule
Model                      & Method                                      & {P} & {R} & {F1} & {BLEU-1} & {ROUGE-L} & {Input Len.} \\ \midrule
\multirow{6}{*}{Qwen3-4B}  & RAG (utterance)                             & 24.58                 & 49.87                 & 28.53                  & 23.20                      & 27.66                       & 75.4                          \\
                           & RAG (observation)                           & 27.25                 & \textbf{51.20}        & 31.17                  & 25.50                      & 30.31                       & 57.1                          \\
                           & Vanilla Training                            & 4.34                  & 13.44                 & 5.76                   & 4.02                       & 5.53                        & 5.6                           \\
                           & \model (no grounding)        & 10.43                 & 21.54                 & 12.26                  & 9.50                       & 11.84                       & 5.6                           \\
                           & \model (static accumulation) & \textbf{29.14}        & 50.42                 & \textbf{32.82}         & \textbf{27.21}             & \textbf{31.96}              & 57.1                          \\
                           & \model (evolving memory)     & 27.95                 & 47.72                 & 31.38                  & 26.23                      & 30.40                       & 51.7                          \\ \midrule
\multirow{6}{*}{Qwen3-8B}  & RAG (utterance)                             & 23.74                 & 48.81                 & 28.10                  & 22.53                      & 27.30                       & 75.4                          \\
                           & RAG (observation)                           & 27.17                 & \textbf{53.52}        & 31.77                  & 25.91                      & 30.63                       & 68.1                          \\
                           & Vanilla Training                            & 4.57                  & 13.22                 & 6.07                   & 4.32                       & 5.73                        & 5.6                           \\
                           & \model (no grounding)        & 10.31                 & 20.37                 & 11.99                  & 9.33                       & 11.67                       & 5.6                           \\
                           & \model (static accumulation) & \textbf{30.99}        & 52.01                 & \textbf{34.50}         & \textbf{29.25}             & \textbf{33.40}              & 68.1                          \\
                           & \model (evolving memory)     & 30.95                 & 50.03                 & 33.99                  & 28.92                      & 32.79                       & 63.2                          \\ \midrule
\multirow{6}{*}{Qwen3-14B} & RAG (utterance)                             & 27.69                 & 49.05                 & 31.38                  & 26.02                      & 30.59                       & 75.4                          \\
                           & RAG (observation)                           & 30.39                 & 54.66                 & 34.83                  & 28.79                      & 33.43                       & 62.4                          \\
                           & Vanilla Training                            & 5.29                  & 12.57                 & 6.59                   & 4.96                       & 6.24                        & 5.6                           \\
                           & \model (no grounding)        & 13.19                 & 22.96                 & 14.71                  & 11.90                      & 14.41                       & 5.6                           \\
                           & \model (static accumulation) & 33.23                 & \textbf{54.98}        & 37.07                  & 31.25                      & 35.65                       & 62.4                          \\
                           & \model (evolving memory)     & {\ul \textbf{34.63}}  & 53.82                 & {\ul \textbf{37.63}}   & {\ul \textbf{32.16}}       & {\ul \textbf{36.16}}        & 61.1                          \\ \midrule
\multirow{6}{*}{Qwen3-32B} & RAG (utterance)                             & 24.31                 & 50.35                 & 28.64                  & 23.19                      & 27.83                       & 75.4                          \\
                           & RAG (observation)                           & 29.20& 56.70                 & 34.00& 28.00& 32.56& 60.1                          \\
                           & Vanilla Training                            & 6.86                  & 15.63                 & 8.28                   & 6.22                       & 7.91                        & 5.6                           \\
                           & \model (no grounding)        & 12.15                 & 26.35                 & 14.81                  & 11.27                      & 14.37                       & 5.6                           \\
                           & \model (static accumulation) & 27.20& {\ul \textbf{57.30}}& 32.75& 26.18& 31.46& 60.1                          \\
                           & \model (evolving memory)     & \textbf{29.58}& 56.09& \textbf{34.25}& \textbf{28.12}& \textbf{32.85}& 59.0                           \\ \bottomrule               
\end{tabular}
}
\caption{Complete results of \model and baselines on LoCoMo dataset (average of three runs), allowing for a holistic understanding of each module's contribution. For each base model, the best results are highlighted in \textbf{bold}, while global optimal performances are {\ul underlined}. The results demonstrate superior information density of observations, usefulness of internalized memory reading, necessity for factual grounding, and efficacy of evolving memory. }
\label{tbl:results_complete}
\end{table*}

%% file: acl_latex.bbl
\begin{thebibliography}{86}
\providecommand{\natexlab}[1]{#1}

\bibitem[{Achiam et~al.(2023)Achiam, Adler, Agarwal, Ahmad, Akkaya, Aleman, Almeida, Altenschmidt, Altman, Anadkat et~al.}]{achiam2023gpt}
Josh Achiam, Steven Adler, Sandhini Agarwal, Lama Ahmad, Ilge Akkaya, Florencia~Leoni Aleman, Diogo Almeida, Janko Altenschmidt, Sam Altman, Shyamal Anadkat, and 1 others. 2023.
\newblock Gpt-4 technical report.
\newblock \emph{arXiv preprint arXiv:2303.08774}.

\bibitem[{Alberti et~al.(2019)Alberti, Andor, Pitler, Devlin, and Collins}]{alberti-etal-2019-synthetic}
Chris Alberti, Daniel Andor, Emily Pitler, Jacob Devlin, and Michael Collins. 2019.
\newblock \href {https://doi.org/10.18653/v1/P19-1620} {Synthetic {QA} corpora generation with roundtrip consistency}.
\newblock In \emph{Proceedings of the 57th Annual Meeting of the Association for Computational Linguistics}, pages 6168--6173, Florence, Italy. Association for Computational Linguistics.

\bibitem[{Amato and Straccia(1999)}]{DBLP:conf/ercimdl/AmatoS99}
Giuseppe Amato and Umberto Straccia. 1999.
\newblock \href {https://doi.org/10.1007/3-540-48155-9\_13} {User profile modeling and applications to digital libraries}.
\newblock In \emph{Research and Advanced Technology for Digital Libraries, Third European Conference, ECDL'99, Paris, France, September 22-24, 1999, Proceedings}, volume 1696 of \emph{Lecture Notes in Computer Science}, pages 184--197. Springer.

\bibitem[{Atkinson and Shiffrin(1968)}]{AtkinsonShiffrin1968}
R.~C. Atkinson and R.~M. Shiffrin. 1968.
\newblock Human memory: A proposed system and its control processes.
\newblock In K.~W. Spence and J.~T. Spence, editors, \emph{The Psychology of Learning and Motivation}, volume~2, pages 89--195. Academic Press, New York.

\bibitem[{Chen et~al.(2025)Chen, Qin, Liu, Peng, Guan, Wang, Hu, Zhou, Gao, and Che}]{chen2025towards}
Qiguang Chen, Libo Qin, Jinhao Liu, Dengyun Peng, Jiannan Guan, Peng Wang, Mengkang Hu, Yuhang Zhou, Te~Gao, and Wanxiang Che. 2025.
\newblock Towards reasoning era: A survey of long chain-of-thought for reasoning large language models.
\newblock \emph{arXiv preprint arXiv:2503.09567}.

\bibitem[{Chhikara et~al.(2025)Chhikara, Khant, Aryan, Singh, and Yadav}]{chhikara2025mem0}
Prateek Chhikara, Dev Khant, Saket Aryan, Taranjeet Singh, and Deshraj Yadav. 2025.
\newblock Mem0: Building production-ready ai agents with scalable long-term memory.
\newblock \emph{arXiv preprint arXiv:2504.19413}.

\bibitem[{Choi et~al.(2023)Choi, Jo, Jang, Jang, and Seo}]{choi-etal-2023-fixed}
Eunbi Choi, Yongrae Jo, Joel Jang, Joonwon Jang, and Minjoon Seo. 2023.
\newblock \href {https://doi.org/10.18653/v1/2023.findings-acl.533} {Fixed input parameterization for efficient prompting}.
\newblock In \emph{Findings of the Association for Computational Linguistics: ACL 2023}, pages 8428--8441, Toronto, Canada. Association for Computational Linguistics.

\bibitem[{Comanici et~al.(2025)Comanici, Bieber, Schaekermann, Pasupat, Sachdeva, Dhillon, Blistein, Ram, Zhang, Rosen et~al.}]{comanici2025gemini}
Gheorghe Comanici, Eric Bieber, Mike Schaekermann, Ice Pasupat, Noveen Sachdeva, Inderjit Dhillon, Marcel Blistein, Ori Ram, Dan Zhang, Evan Rosen, and 1 others. 2025.
\newblock Gemini 2.5: Pushing the frontier with advanced reasoning, multimodality, long context, and next generation agentic capabilities.
\newblock \emph{arXiv preprint arXiv:2507.06261}.

\bibitem[{Cover(1999)}]{cover1999elements}
Thomas~M Cover. 1999.
\newblock \emph{Elements of information theory}.
\newblock John Wiley \& Sons.

\bibitem[{Deng et~al.(2023)Deng, Zhang, Liu, Wu, Wang, and Mihalcea}]{deng-etal-2023-annotate}
Naihao Deng, Xinliang Zhang, Siyang Liu, Winston Wu, Lu~Wang, and Rada Mihalcea. 2023.
\newblock \href {https://doi.org/10.18653/v1/2023.findings-emnlp.832} {You are what you annotate: Towards better models through annotator representations}.
\newblock In \emph{Findings of the Association for Computational Linguistics: EMNLP 2023}, pages 12475--12498, Singapore. Association for Computational Linguistics.

\bibitem[{Devlin et~al.(2019)Devlin, Chang, Lee, and Toutanova}]{devlin-etal-2019-bert}
Jacob Devlin, Ming-Wei Chang, Kenton Lee, and Kristina Toutanova. 2019.
\newblock \href {https://doi.org/10.18653/v1/N19-1423} {{BERT}: Pre-training of deep bidirectional transformers for language understanding}.
\newblock In \emph{Proceedings of the 2019 Conference of the North {A}merican Chapter of the Association for Computational Linguistics: Human Language Technologies, Volume 1 (Long and Short Papers)}, pages 4171--4186, Minneapolis, Minnesota. Association for Computational Linguistics.

\bibitem[{Douze et~al.(2024)Douze, Guzhva, Deng, Johnson, Szilvasy, Mazaré, Lomeli, Hosseini, and Jégou}]{douze2024faiss}
Matthijs Douze, Alexandr Guzhva, Chengqi Deng, Jeff Johnson, Gergely Szilvasy, Pierre-Emmanuel Mazaré, Maria Lomeli, Lucas Hosseini, and Hervé Jégou. 2024.
\newblock \href {https://arxiv.org/abs/2401.08281} {The faiss library}.

\bibitem[{Du et~al.(2025)Du, Huang, Zheng, Wang, Montella, Lapata, Wong, and Pan}]{du2025rethinking}
Yiming Du, Wenyu Huang, Danna Zheng, Zhaowei Wang, Sebastien Montella, Mirella Lapata, Kam-Fai Wong, and Jeff~Z Pan. 2025.
\newblock Rethinking memory in ai: Taxonomy, operations, topics, and future directions.
\newblock \emph{arXiv e-prints}, pages arXiv--2505.

\bibitem[{Duran and Aytekin(2025)}]{duran2025beyond}
Mehmet~Samet Duran and Tevfik Aytekin. 2025.
\newblock Beyond one-size-fits-all summarization: Customizing summaries for diverse users.
\newblock \emph{arXiv preprint arXiv:2503.10675}.

\bibitem[{Fareed et~al.(2023)Fareed, Hassan, Belhaouari, and Halim}]{FAREED2023100495}
Aamir Fareed, Saima Hassan, Samir~Brahim Belhaouari, and Zahid Halim. 2023.
\newblock \href {https://doi.org/10.1016/j.mlwa.2023.100495} {A collaborative filtering recommendation framework utilizing social networks}.
\newblock \emph{Machine Learning with Applications}, 14:100495.

\bibitem[{Fischer(2001)}]{fischer2001user}
Gerhard Fischer. 2001.
\newblock User modeling in human--computer interaction.
\newblock \emph{User modeling and user-adapted interaction}, 11(1):65--86.

\bibitem[{Gao et~al.(2013)Gao, Hao, Bai, Li, Li, and Zhu}]{DBLP:conf/recsys/GaoHBLLZ13}
Rui Gao, Bibo Hao, Shuotian Bai, Lin Li, Ang Li, and Tingshao Zhu. 2013.
\newblock \href {https://doi.org/10.1145/2507157.2507219} {Improving user profile with personality traits predicted from social media content}.
\newblock In \emph{Seventh {ACM} Conference on Recommender Systems, RecSys '13, Hong Kong, China, October 12-16, 2013}, pages 355--358. {ACM}.

\bibitem[{Gou et~al.(2014)Gou, Zhou, and Yang}]{DBLP:conf/chi/GouZY14}
Liang Gou, Michelle~X. Zhou, and Huahai Yang. 2014.
\newblock \href {https://doi.org/10.1145/2556288.2557398} {Knowme and shareme: understanding automatically discovered personality traits from social media and user sharing preferences}.
\newblock In \emph{{CHI} Conference on Human Factors in Computing Systems, CHI'14, Toronto, ON, Canada - April 26 - May 01, 2014}, pages 955--964. {ACM}.

\bibitem[{He et~al.(2017)He, Liao, Zhang, Nie, Hu, and Chua}]{he2017neural}
Xiangnan He, Lizi Liao, Hanwang Zhang, Liqiang Nie, Xia Hu, and Tat-Seng Chua. 2017.
\newblock Neural collaborative filtering.
\newblock In \emph{Proceedings of the 26th international conference on world wide web}, pages 173--182.

\bibitem[{Hoch and Loewenstein(1991)}]{hoch1991time}
Stephen~J Hoch and George~F Loewenstein. 1991.
\newblock Time-inconsistent preferences and consumer self-control.
\newblock \emph{Journal of consumer research}, 17(4):492--507.

\bibitem[{Hu et~al.(2022)Hu, Shen, Wallis, Allen-Zhu, Li, Wang, Wang, Chen et~al.}]{hu2022lora}
Edward~J Hu, Yelong Shen, Phillip Wallis, Zeyuan Allen-Zhu, Yuanzhi Li, Shean Wang, Lu~Wang, Weizhu Chen, and 1 others. 2022.
\newblock Lora: Low-rank adaptation of large language models.
\newblock \emph{ICLR}, 1(2):3.

\bibitem[{Hughes et~al.(2020)Hughes, Agarwal, Guo, and Sycara}]{hughes2020inferring}
Dana Hughes, Akshat Agarwal, Yue Guo, and Katia Sycara. 2020.
\newblock Inferring non-stationary human preferences for human-agent teams.
\newblock In \emph{2020 29th IEEE International Conference on Robot and Human Interactive Communication (RO-MAN)}, pages 1178--1185. IEEE.

\bibitem[{Jiang et~al.(2014)Jiang, Cui, Wang, Zhu, and Yang}]{DBLP:journals/tkde/JiangCWZY14}
Meng Jiang, Peng Cui, Fei Wang, Wenwu Zhu, and Shiqiang Yang. 2014.
\newblock \href {https://doi.org/10.1109/TKDE.2014.2300487} {Scalable recommendation with social contextual information}.
\newblock \emph{{IEEE} Trans. Knowl. Data Eng.}, 26(11):2789--2802.

\bibitem[{Kang et~al.(2023)Kang, Ni, Mehta, Sathiamoorthy, Hong, Chi, and Cheng}]{DBLP:journals/corr/abs-2305-06474}
Wang{-}Cheng Kang, Jianmo Ni, Nikhil Mehta, Maheswaran Sathiamoorthy, Lichan Hong, Ed~H. Chi, and Derek~Zhiyuan Cheng. 2023.
\newblock \href {https://doi.org/10.48550/ARXIV.2305.06474} {Do llms understand user preferences? evaluating llms on user rating prediction}.
\newblock \emph{CoRR}, abs/2305.06474.

\bibitem[{Kim et~al.(2013)Kim, Lee, and Ryu}]{kim2013personality}
Jieun Kim, Ahreum Lee, and Hokyoung Ryu. 2013.
\newblock Personality and its effects on learning performance: Design guidelines for an adaptive e-learning system based on a user model.
\newblock \emph{International Journal of Industrial Ergonomics}, 43(5):450--461.

\bibitem[{Kim et~al.(2025)Kim, Lee, Kim, Kim, and Cho}]{kim2025pre}
Sangyeop Kim, Yohan Lee, Sanghwa Kim, Hyunjong Kim, and Sungzoon Cho. 2025.
\newblock Pre-storage reasoning for episodic memory: Shifting inference burden to memory for personalized dialogue.
\newblock \emph{arXiv preprint arXiv:2509.10852}.

\bibitem[{Koren et~al.(2009)Koren, Bell, and Volinsky}]{DBLP:journals/computer/KorenBV09}
Yehuda Koren, Robert~M. Bell, and Chris Volinsky. 2009.
\newblock \href {https://doi.org/10.1109/MC.2009.263} {Matrix factorization techniques for recommender systems}.
\newblock \emph{Computer}, 42(8):30--37.

\bibitem[{Lee et~al.(2025)Lee, Sakaguchi, and Bak}]{lee-etal-2025-self}
Jaehyeok Lee, Keisuke Sakaguchi, and JinYeong Bak. 2025.
\newblock \href {https://doi.org/10.18653/v1/2025.naacl-long.528} {Self-training meets consistency: Improving {LLM}s' reasoning with consistency-driven rationale evaluation}.
\newblock In \emph{Proceedings of the 2025 Conference of the Nations of the Americas Chapter of the Association for Computational Linguistics: Human Language Technologies (Volume 1: Long Papers)}, pages 10519--10539, Albuquerque, New Mexico. Association for Computational Linguistics.

\bibitem[{Lewis et~al.(2020)Lewis, Perez, Piktus, Petroni, Karpukhin, Goyal, K{\"u}ttler, Lewis, Yih, Rockt{\"a}schel et~al.}]{lewis2020retrieval}
Patrick Lewis, Ethan Perez, Aleksandra Piktus, Fabio Petroni, Vladimir Karpukhin, Naman Goyal, Heinrich K{\"u}ttler, Mike Lewis, Wen-tau Yih, Tim Rockt{\"a}schel, and 1 others. 2020.
\newblock Retrieval-augmented generation for knowledge-intensive nlp tasks.
\newblock \emph{Advances in neural information processing systems}, 33:9459--9474.

\bibitem[{Li et~al.(2025)Li, Zhou, and Song}]{li-etal-2025-bild}
Minchong Li, Feng Zhou, and Xiaohui Song. 2025.
\newblock \href {https://aclanthology.org/2025.coling-main.78/} {{B}i{LD}: Bi-directional logits difference loss for large language model distillation}.
\newblock In \emph{Proceedings of the 31st International Conference on Computational Linguistics}, pages 1168--1182, Abu Dhabi, UAE. Association for Computational Linguistics.

\bibitem[{Liu et~al.(2025)Liu, Qiu, Li, Dai, Yu, Zhu, Hu, Yang, Chua, and King}]{liu2025survey}
Jiahong Liu, Zexuan Qiu, Zhongyang Li, Quanyu Dai, Wenhao Yu, Jieming Zhu, Minda Hu, Menglin Yang, Tat-Seng Chua, and Irwin King. 2025.
\newblock A survey of personalized large language models: Progress and future directions.
\newblock \emph{arXiv preprint arXiv:2502.11528}.

\bibitem[{Liu et~al.(2023)Liu, Liu, Lv, Zhou, and Zhang}]{DBLP:journals/corr/abs-2304-10149}
Junling Liu, Chao Liu, Renjie Lv, Kang Zhou, and Yan Zhang. 2023.
\newblock \href {https://doi.org/10.48550/ARXIV.2304.10149} {Is chatgpt a good recommender? {A} preliminary study}.
\newblock \emph{CoRR}, abs/2304.10149.

\bibitem[{Liu et~al.(2024)Liu, Lin, Hewitt, Paranjape, Bevilacqua, Petroni, and Liang}]{liu-etal-2024-lost}
Nelson~F. Liu, Kevin Lin, John Hewitt, Ashwin Paranjape, Michele Bevilacqua, Fabio Petroni, and Percy Liang. 2024.
\newblock \href {https://doi.org/10.1162/tacl_a_00638} {Lost in the middle: How language models use long contexts}.
\newblock \emph{Transactions of the Association for Computational Linguistics}, 12:157--173.

\bibitem[{Luo et~al.(2026)Luo, Tian, Cao, Luo, Lin, Li, Kong, Yang, and Ma}]{luo2026storage}
Jinghao Luo, Yuchen Tian, Chuxue Cao, Ziyang Luo, Hongzhan Lin, Kaixin Li, Chuyi Kong, Ruichao Yang, and Jing Ma. 2026.
\newblock From storage to experience: A survey on the evolution of llm agent memory mechanisms.

\bibitem[{Madaan et~al.(2022)Madaan, Tandon, Clark, and Yang}]{madaan-etal-2022-memory}
Aman Madaan, Niket Tandon, Peter Clark, and Yiming Yang. 2022.
\newblock \href {https://doi.org/10.18653/v1/2022.emnlp-main.183} {Memory-assisted prompt editing to improve {GPT}-3 after deployment}.
\newblock In \emph{Proceedings of the 2022 Conference on Empirical Methods in Natural Language Processing}, pages 2833--2861, Abu Dhabi, United Arab Emirates. Association for Computational Linguistics.

\bibitem[{Magister et~al.(2024)Magister, Metcalf, Zhang, and ter Hoeve}]{DBLP:journals/corr/abs-2411-13405}
Lucie~Charlotte Magister, Katherine Metcalf, Yizhe Zhang, and Maartje ter Hoeve. 2024.
\newblock \href {https://doi.org/10.48550/ARXIV.2411.13405} {On the way to {LLM} personalization: Learning to remember user conversations}.
\newblock \emph{CoRR}, abs/2411.13405.

\bibitem[{Magister et~al.(2025)Magister, Metcalf, Zhang, and Ter~Hoeve}]{magister-etal-2025-way}
Lucie~Charlotte Magister, Katherine Metcalf, Yizhe Zhang, and Maartje Ter~Hoeve. 2025.
\newblock \href {https://doi.org/10.18653/v1/2025.l2m2-1.5} {On the way to {LLM} personalization: Learning to remember user conversations}.
\newblock In \emph{Proceedings of the First Workshop on Large Language Model Memorization (L2M2)}, pages 61--77, Vienna, Austria. Association for Computational Linguistics.

\bibitem[{Maharana et~al.(2024)Maharana, Lee, Tulyakov, Bansal, Barbieri, and Fang}]{maharana-etal-2024-evaluating}
Adyasha Maharana, Dong-Ho Lee, Sergey Tulyakov, Mohit Bansal, Francesco Barbieri, and Yuwei Fang. 2024.
\newblock \href {https://doi.org/10.18653/v1/2024.acl-long.747} {Evaluating very long-term conversational memory of {LLM} agents}.
\newblock In \emph{Proceedings of the 62nd Annual Meeting of the Association for Computational Linguistics (Volume 1: Long Papers)}, pages 13851--13870, Bangkok, Thailand. Association for Computational Linguistics.

\bibitem[{Mysore et~al.(2024)Mysore, Lu, Wan, Yang, Sarrafzadeh, Menezes, Baghaee, Gonzalez, Neville, and Safavi}]{mysore-etal-2024-pearl}
Sheshera Mysore, Zhuoran Lu, Mengting Wan, Longqi Yang, Bahareh Sarrafzadeh, Steve Menezes, Tina Baghaee, Emmanuel~Barajas Gonzalez, Jennifer Neville, and Tara Safavi. 2024.
\newblock \href {https://doi.org/10.18653/v1/2024.customnlp4u-1.16} {Pearl: Personalizing large language model writing assistants with generation-calibrated retrievers}.
\newblock In \emph{Proceedings of the 1st Workshop on Customizable NLP: Progress and Challenges in Customizing NLP for a Domain, Application, Group, or Individual (CustomNLP4U)}, pages 198--219, Miami, Florida, USA. Association for Computational Linguistics.

\bibitem[{Nan et~al.(2025)Nan, Ma, Wu, and Chen}]{nan2025nemori}
Jiayan Nan, Wenquan Ma, Wenlong Wu, and Yize Chen. 2025.
\newblock Nemori: Self-organizing agent memory inspired by cognitive science.
\newblock \emph{arXiv preprint arXiv:2508.03341}.

\bibitem[{Park et~al.(2023)Park, O'Brien, Cai, Morris, Liang, and Bernstein}]{park2023generative}
Joon~Sung Park, Joseph O'Brien, Carrie~Jun Cai, Meredith~Ringel Morris, Percy Liang, and Michael~S Bernstein. 2023.
\newblock Generative agents: Interactive simulacra of human behavior.
\newblock In \emph{Proceedings of the 36th annual acm symposium on user interface software and technology}, pages 1--22.

\bibitem[{Petrov and Macdonald(2023)}]{DBLP:journals/corr/abs-2306-11114}
Aleksandr~V. Petrov and Craig Macdonald. 2023.
\newblock \href {https://doi.org/10.48550/ARXIV.2306.11114} {Generative sequential recommendation with gptrec}.
\newblock \emph{CoRR}, abs/2306.11114.

\bibitem[{Pham et~al.(2022)Pham, Cho, Joshi, and Hegde}]{pham2022revisiting}
Minh Pham, Minsu Cho, Ameya Joshi, and Chinmay Hegde. 2022.
\newblock Revisiting self-distillation.
\newblock \emph{arXiv preprint arXiv:2206.08491}.

\bibitem[{Piryani et~al.(2025)Piryani, Abdullah, Mozafari, Anand, and Jatowt}]{piryani2025s}
Bhawna Piryani, Abdelrahman Abdullah, Jamshid Mozafari, Avishek Anand, and Adam Jatowt. 2025.
\newblock It's high time: A survey of temporal information retrieval and question answering.
\newblock \emph{arXiv e-prints}, pages arXiv--2505.

\bibitem[{Purificato et~al.(2024)Purificato, Boratto, and Luca}]{DBLP:journals/corr/abs-2402-09660}
Erasmo Purificato, Ludovico Boratto, and Ernesto William~De Luca. 2024.
\newblock \href {https://doi.org/10.48550/ARXIV.2402.09660} {User modeling and user profiling: {A} comprehensive survey}.
\newblock \emph{CoRR}, abs/2402.09660.

\bibitem[{Qin et~al.(2024)Qin, Xia, Jia, Jiang, Abbasi, Zhou, Hu, and Shi}]{qin2024enabling}
Ruiyang Qin, Jun Xia, Zhenge Jia, Meng Jiang, Ahmed Abbasi, Peipei Zhou, Jingtong Hu, and Yiyu Shi. 2024.
\newblock Enabling on-device large language model personalization with self-supervised data selection and synthesis.
\newblock In \emph{Proceedings of the 61st ACM/IEEE design automation conference}, pages 1--6.

\bibitem[{Qiu et~al.(2025)Qiu, Zhao, Zhang, Bai, Wang, Cheng, Feng, and Chua}]{qiu-etal-2025-measuring}
Yilun Qiu, Xiaoyan Zhao, Yang Zhang, Yimeng Bai, Wenjie Wang, Hong Cheng, Fuli Feng, and Tat-Seng Chua. 2025.
\newblock \href {https://doi.org/10.18653/v1/2025.findings-acl.1095} {Measuring what makes you unique: Difference-aware user modeling for enhancing {LLM} personalization}.
\newblock In \emph{Findings of the Association for Computational Linguistics: ACL 2025}, pages 21258--21277, Vienna, Austria. Association for Computational Linguistics.

\bibitem[{Qiu et~al.(2021)Qiu, Wu, Gao, and Fan}]{DBLP:conf/aaai/QiuWG021}
Zhaopeng Qiu, Xian Wu, Jingyue Gao, and Wei Fan. 2021.
\newblock \href {https://doi.org/10.1609/AAAI.V35I5.16557} {{U-BERT:} pre-training user representations for improved recommendation}.
\newblock In \emph{Thirty-Fifth {AAAI} Conference on Artificial Intelligence, {AAAI} 2021, Thirty-Third Conference on Innovative Applications of Artificial Intelligence, {IAAI} 2021, The Eleventh Symposium on Educational Advances in Artificial Intelligence, {EAAI} 2021, Virtual Event, February 2-9, 2021}, pages 4320--4327. {AAAI} Press.

\bibitem[{Rasmussen et~al.(2025)Rasmussen, Paliychuk, Beauvais, Ryan, and Chalef}]{rasmussen2025zep}
Preston Rasmussen, Pavlo Paliychuk, Travis Beauvais, Jack Ryan, and Daniel Chalef. 2025.
\newblock Zep: a temporal knowledge graph architecture for agent memory.
\newblock \emph{arXiv preprint arXiv:2501.13956}.

\bibitem[{Richardson et~al.(2023)Richardson, Zhang, Gillespie, Kar, Singh, Raeesy, Khan, and Sethy}]{DBLP:journals/corr/abs-2310-20081}
Christopher Richardson, Yao Zhang, Kellen Gillespie, Sudipta Kar, Arshdeep Singh, Zeynab Raeesy, Omar~Zia Khan, and Abhinav Sethy. 2023.
\newblock \href {https://doi.org/10.48550/ARXIV.2310.20081} {Integrating summarization and retrieval for enhanced personalization via large language models}.
\newblock \emph{CoRR}, abs/2310.20081.

\bibitem[{Risko and Gilbert(2016)}]{risko2016cognitive}
Evan~F Risko and Sam~J Gilbert. 2016.
\newblock Cognitive offloading.
\newblock \emph{Trends in cognitive sciences}, 20(9):676--688.

\bibitem[{Robertson and Zaragoza(2009)}]{DBLP:journals/ftir/RobertsonZ09}
Stephen~E. Robertson and Hugo Zaragoza. 2009.
\newblock \href {https://doi.org/10.1561/1500000019} {The probabilistic relevance framework: {BM25} and beyond}.
\newblock \emph{Found. Trends Inf. Retr.}, 3(4):333--389.

\bibitem[{Salama et~al.(2025)Salama, Cai, Yuan, Currey, Sunkara, Zhang, and Benajiba}]{salama-etal-2025-meminsight}
Rana Salama, Jason Cai, Michelle Yuan, Anna Currey, Monica Sunkara, Yi~Zhang, and Yassine Benajiba. 2025.
\newblock \href {https://doi.org/10.18653/v1/2025.emnlp-main.1683} {{M}em{I}nsight: Autonomous memory augmentation for {LLM} agents}.
\newblock In \emph{Proceedings of the 2025 Conference on Empirical Methods in Natural Language Processing}, pages 33136--33152, Suzhou, China. Association for Computational Linguistics.

\bibitem[{Salemi et~al.(2024)Salemi, Mysore, Bendersky, and Zamani}]{salemi-etal-2024-lamp}
Alireza Salemi, Sheshera Mysore, Michael Bendersky, and Hamed Zamani. 2024.
\newblock \href {https://doi.org/10.18653/v1/2024.acl-long.399} {{L}a{MP}: When large language models meet personalization}.
\newblock In \emph{Proceedings of the 62nd Annual Meeting of the Association for Computational Linguistics (Volume 1: Long Papers)}, pages 7370--7392, Bangkok, Thailand. Association for Computational Linguistics.

\bibitem[{Schulman et~al.(2017)Schulman, Wolski, Dhariwal, Radford, and Klimov}]{schulman2017proximal}
John Schulman, Filip Wolski, Prafulla Dhariwal, Alec Radford, and Oleg Klimov. 2017.
\newblock Proximal policy optimization algorithms.
\newblock \emph{arXiv preprint arXiv:1707.06347}.

\bibitem[{Shao et~al.(2024)Shao, Wang, Zhu, Xu, Song, Bi, Zhang, Zhang, Li, Wu et~al.}]{shao2024deepseekmath}
Zhihong Shao, Peiyi Wang, Qihao Zhu, Runxin Xu, Junxiao Song, Xiao Bi, Haowei Zhang, Mingchuan Zhang, YK~Li, Yang Wu, and 1 others. 2024.
\newblock Deepseekmath: Pushing the limits of mathematical reasoning in open language models.
\newblock \emph{arXiv preprint arXiv:2402.03300}.

\bibitem[{Singh et~al.(2025)Singh, Fry, Perelman, Tart, Ganesh, El-Kishky, McLaughlin, Low, Ostrow, Ananthram et~al.}]{singh2025openai}
Aaditya Singh, Adam Fry, Adam Perelman, Adam Tart, Adi Ganesh, Ahmed El-Kishky, Aidan McLaughlin, Aiden Low, AJ~Ostrow, Akhila Ananthram, and 1 others. 2025.
\newblock Openai gpt-5 system card.
\newblock \emph{arXiv preprint arXiv:2601.03267}.

\bibitem[{Snell et~al.(2022)Snell, Klein, and Zhong}]{snell2022learning}
Charlie Snell, Dan Klein, and Ruiqi Zhong. 2022.
\newblock Learning by distilling context.
\newblock \emph{arXiv preprint arXiv:2209.15189}.

\bibitem[{Sweller(1988)}]{sweller1988cognitive}
John Sweller. 1988.
\newblock Cognitive load during problem solving: Effects on learning.
\newblock \emph{Cognitive science}, 12(2):257--285.

\bibitem[{Tan et~al.(2025)Tan, Zhang, Ma, Chen, Dai, and Dong}]{tan-etal-2025-membench}
Haoran Tan, Zeyu Zhang, Chen Ma, Xu~Chen, Quanyu Dai, and Zhenhua Dong. 2025.
\newblock \href {https://doi.org/10.18653/v1/2025.findings-acl.989} {{M}em{B}ench: Towards more comprehensive evaluation on the memory of {LLM}-based agents}.
\newblock In \emph{Findings of the Association for Computational Linguistics: ACL 2025}, pages 19336--19352, Vienna, Austria. Association for Computational Linguistics.

\bibitem[{Tan et~al.(2023)Tan, Ng, and Bing}]{tan-etal-2023-towards}
Qingyu Tan, Hwee~Tou Ng, and Lidong Bing. 2023.
\newblock \href {https://doi.org/10.18653/v1/2023.acl-long.828} {Towards benchmarking and improving the temporal reasoning capability of large language models}.
\newblock In \emph{Proceedings of the 61st Annual Meeting of the Association for Computational Linguistics (Volume 1: Long Papers)}, pages 14820--14835, Toronto, Canada. Association for Computational Linguistics.

\bibitem[{Tan et~al.(2024{\natexlab{a}})Tan, Liu, and Jiang}]{tan-etal-2024-personalized}
Zhaoxuan Tan, Zheyuan Liu, and Meng Jiang. 2024{\natexlab{a}}.
\newblock \href {https://doi.org/10.18653/v1/2024.emnlp-main.371} {Personalized pieces: Efficient personalized large language models through collaborative efforts}.
\newblock In \emph{Proceedings of the 2024 Conference on Empirical Methods in Natural Language Processing}, pages 6459--6475, Miami, Florida, USA. Association for Computational Linguistics.

\bibitem[{Tan et~al.(2024{\natexlab{b}})Tan, Zeng, Tian, Liu, Yin, and Jiang}]{tan-etal-2024-democratizing}
Zhaoxuan Tan, Qingkai Zeng, Yijun Tian, Zheyuan Liu, Bing Yin, and Meng Jiang. 2024{\natexlab{b}}.
\newblock \href {https://doi.org/10.18653/v1/2024.emnlp-main.372} {Democratizing large language models via personalized parameter-efficient fine-tuning}.
\newblock In \emph{Proceedings of the 2024 Conference on Empirical Methods in Natural Language Processing}, pages 6476--6491, Miami, Florida, USA. Association for Computational Linguistics.

\bibitem[{Tang et~al.(2024)Tang, Zou, Zhang, Li, Zhao, Zhang, Cohan, and Gerstein}]{tang-etal-2024-medagents}
Xiangru Tang, Anni Zou, Zhuosheng Zhang, Ziming Li, Yilun Zhao, Xingyao Zhang, Arman Cohan, and Mark Gerstein. 2024.
\newblock \href {https://doi.org/10.18653/v1/2024.findings-acl.33} {{M}ed{A}gents: Large language models as collaborators for zero-shot medical reasoning}.
\newblock In \emph{Findings of the Association for Computational Linguistics: ACL 2024}, pages 599--621, Bangkok, Thailand. Association for Computational Linguistics.

\bibitem[{Tseng et~al.(2024)Tseng, Huang, Hsiao, Chen, Huang, Meng, and Chen}]{tseng-etal-2024-two}
Yu-Min Tseng, Yu-Chao Huang, Teng-Yun Hsiao, Wei-Lin Chen, Chao-Wei Huang, Yu~Meng, and Yun-Nung Chen. 2024.
\newblock \href {https://doi.org/10.18653/v1/2024.findings-emnlp.969} {Two tales of persona in {LLM}s: A survey of role-playing and personalization}.
\newblock In \emph{Findings of the Association for Computational Linguistics: EMNLP 2024}, pages 16612--16631, Miami, Florida, USA. Association for Computational Linguistics.

\bibitem[{Tulving(1985)}]{tulving1985many}
Endel Tulving. 1985.
\newblock How many memory systems are there?
\newblock \emph{American psychologist}, 40(4):385.

\bibitem[{Tulving et~al.(1972)}]{tulving1972episodic}
Endel Tulving and 1 others. 1972.
\newblock Episodic and semantic memory.
\newblock \emph{Organization of memory}, 1(381-403):1.

\bibitem[{Vaswani et~al.(2017)Vaswani, Shazeer, Parmar, Uszkoreit, Jones, Gomez, Kaiser, and Polosukhin}]{transformer}
Ashish Vaswani, Noam Shazeer, Niki Parmar, Jakob Uszkoreit, Llion Jones, Aidan~N. Gomez, Lukasz Kaiser, and Illia Polosukhin. 2017.
\newblock \href {https://proceedings.neurips.cc/paper/2017/hash/3f5ee243547dee91fbd053c1c4a845aa-Abstract.html} {Attention is all you need}.
\newblock In \emph{Advances in Neural Information Processing Systems 30: Annual Conference on Neural Information Processing Systems 2017, December 4-9, 2017, Long Beach, CA, {USA}}, pages 5998--6008.

\bibitem[{Wang et~al.(2025{\natexlab{a}})Wang, Yang, Yin, and Gao}]{wang2025never}
Haoming Wang, Boyuan Yang, Xiangyu Yin, and Wei Gao. 2025{\natexlab{a}}.
\newblock Never start from scratch: Expediting on-device llm personalization via explainable model selection.
\newblock In \emph{Proceedings of the 23rd Annual International Conference on Mobile Systems, Applications and Services}, pages 154--168.

\bibitem[{Wang et~al.(2024)Wang, Ma, Feng, Zhang, Yang, Zhang, Chen, Tang, Chen, Lin et~al.}]{wang2024survey}
Lei Wang, Chen Ma, Xueyang Feng, Zeyu Zhang, Hao Yang, Jingsen Zhang, Zhiyuan Chen, Jiakai Tang, Xu~Chen, Yankai Lin, and 1 others. 2024.
\newblock A survey on large language model based autonomous agents.
\newblock \emph{Frontiers of Computer Science}, 18(6):186345.

\bibitem[{Wang et~al.(2025{\natexlab{b}})Wang, Takanobu, Liang, Mao, Hu, McAuley, and Wu}]{wang2025mem}
Yu~Wang, Ryuichi Takanobu, Zhiqi Liang, Yuzhen Mao, Yuanzhe Hu, Julian McAuley, and Xiaojian Wu. 2025{\natexlab{b}}.
\newblock Mem-$\{$$\backslash$alpha$\}$: Learning memory construction via reinforcement learning.
\newblock \emph{arXiv preprint arXiv:2509.25911}.

\bibitem[{Wei et~al.(2025)Wei, Sachdeva, Coleman, He, Bei, Ning, Ai, Li, He, Chi et~al.}]{wei2025evo}
Tianxin Wei, Noveen Sachdeva, Benjamin Coleman, Zhankui He, Yuanchen Bei, Xuying Ning, Mengting Ai, Yunzhe Li, Jingrui He, Ed~H Chi, and 1 others. 2025.
\newblock Evo-memory: Benchmarking llm agent test-time learning with self-evolving memory.
\newblock \emph{arXiv preprint arXiv:2511.20857}.

\bibitem[{Wu et~al.(2025{\natexlab{a}})Wu, Wang, Yu, Zhang, Chang, and Yu}]{DBLP:conf/iclr/WuWYZCY25}
Di~Wu, Hongwei Wang, Wenhao Yu, Yuwei Zhang, Kai{-}Wei Chang, and Dong Yu. 2025{\natexlab{a}}.
\newblock \href {https://openreview.net/forum?id=pZiyCaVuti} {Longmemeval: Benchmarking chat assistants on long-term interactive memory}.
\newblock In \emph{The Thirteenth International Conference on Learning Representations, {ICLR} 2025, Singapore, April 24-28, 2025}. OpenReview.net.

\bibitem[{Wu et~al.(2025{\natexlab{b}})Wu, Liang, Zhang, Wang, Zhang, Guo, Tang, and Liu}]{wu2025human}
Yaxiong Wu, Sheng Liang, Chen Zhang, Yichao Wang, Yongyue Zhang, Huifeng Guo, Ruiming Tang, and Yong Liu. 2025{\natexlab{b}}.
\newblock From human memory to ai memory: A survey on memory mechanisms in the era of llms.
\newblock \emph{arXiv preprint arXiv:2504.15965}.

\bibitem[{Xu et~al.(2025)Xu, Liang, Mei, Gao, Tan, and Zhang}]{xu2025mem}
Wujiang Xu, Zujie Liang, Kai Mei, Hang Gao, Juntao Tan, and Yongfeng Zhang. 2025.
\newblock A-mem: Agentic memory for llm agents.
\newblock \emph{arXiv preprint arXiv:2502.12110}.

\bibitem[{Yan et~al.(2025)Yan, Yang, Huang, Nie, Ding, Li, Ma, Bi, Kersting, Pan, Schuetze, Tresp, and Ma}]{yan2025memory}
Sikuan Yan, Xiufeng Yang, Zuchao Huang, Ercong Nie, Zifeng Ding, Zonggen Li, Xiaowen Ma, Jinhe Bi, Kristian Kersting, Jeff~Z. Pan, Hinrich Schuetze, Volker Tresp, and Yunpu Ma. 2025.
\newblock {Memory-R1}: Enhancing large language model agents to manage and utilize memories via reinforcement learning.
\newblock \emph{arXiv preprint arXiv:2508.19828}.

\bibitem[{Yang et~al.(2025)Yang, Li, Yang, Zhang, Hui, Zheng, Yu, Gao, Huang, Lv et~al.}]{yang2025qwen3}
An~Yang, Anfeng Li, Baosong Yang, Beichen Zhang, Binyuan Hui, Bo~Zheng, Bowen Yu, Chang Gao, Chengen Huang, Chenxu Lv, and 1 others. 2025.
\newblock Qwen3 technical report.
\newblock \emph{arXiv preprint arXiv:2505.09388}.

\bibitem[{Yang et~al.(2024)Yang, Yang, Zhang, Hui, Zheng, Yu, Li, Liu, Huang, Wei, Lin, Yang, Tu, Zhang, Yang, Yang, Zhou, Lin, Dang, Lu, Bao, Yang, Yu, Li, Xue, Zhang, Zhu, Men, Lin, Li, Xia, Ren, Ren, Fan, Su, Zhang, Wan, Liu, Cui, Zhang, and Qiu}]{DBLP:journals/corr/abs-2412-15115}
An~Yang, Baosong Yang, Beichen Zhang, Binyuan Hui, Bo~Zheng, Bowen Yu, Chengyuan Li, Dayiheng Liu, Fei Huang, Haoran Wei, Huan Lin, Jian Yang, Jianhong Tu, Jianwei Zhang, Jianxin Yang, Jiaxi Yang, Jingren Zhou, Junyang Lin, Kai Dang, and 22 others. 2024.
\newblock \href {https://doi.org/10.48550/ARXIV.2412.15115} {Qwen2.5 technical report}.
\newblock \emph{CoRR}, abs/2412.15115.

\bibitem[{Zhang et~al.(2024{\natexlab{a}})Zhang, Kang, Zhao, and Liu}]{zhang-etal-2024-llm-based}
Kai Zhang, Yangyang Kang, Fubang Zhao, and Xiaozhong Liu. 2024{\natexlab{a}}.
\newblock \href {https://doi.org/10.18653/v1/2024.naacl-long.132} {{LLM}-based medical assistant personalization with short- and long-term memory coordination}.
\newblock In \emph{Proceedings of the 2024 Conference of the North American Chapter of the Association for Computational Linguistics: Human Language Technologies (Volume 1: Long Papers)}, pages 2386--2398, Mexico City, Mexico. Association for Computational Linguistics.

\bibitem[{Zhang et~al.(2024{\natexlab{b}})Zhang, Qing, Kang, and Liu}]{DBLP:journals/corr/abs-2404-03565}
Kai Zhang, Lizhi Qing, Yangyang Kang, and Xiaozhong Liu. 2024{\natexlab{b}}.
\newblock \href {https://doi.org/10.48550/ARXIV.2404.03565} {Personalized {LLM} response generation with parameterized memory injection}.
\newblock \emph{CoRR}, abs/2404.03565.

\bibitem[{Zhang et~al.(2019)Zhang, Song, Gao, Chen, Bao, and Ma}]{zhang2019your}
Linfeng Zhang, Jiebo Song, Anni Gao, Jingwei Chen, Chenglong Bao, and Kaisheng Ma. 2019.
\newblock Be your own teacher: Improve the performance of convolutional neural networks via self distillation.
\newblock In \emph{Proceedings of the IEEE/CVF international conference on computer vision}, pages 3713--3722.

\bibitem[{Zhang et~al.(2025)Zhang, Beauchamp, and Wang}]{zhang-etal-2025-prime}
Xinliang~Frederick Zhang, Nicholas Beauchamp, and Lu~Wang. 2025.
\newblock \href {https://doi.org/10.18653/v1/2025.emnlp-main.1711} {{PRIME}: Large language model personalization with cognitive dual-memory and personalized thought process}.
\newblock In \emph{Proceedings of the 2025 Conference on Empirical Methods in Natural Language Processing}, pages 33707--33736, Suzhou, China. Association for Computational Linguistics.

\bibitem[{Zhang et~al.(2024{\natexlab{c}})Zhang, Beauchamp, and Wang}]{zhang-etal-2024-narrative}
Xinliang~Frederick Zhang, Nick Beauchamp, and Lu~Wang. 2024{\natexlab{c}}.
\newblock \href {https://doi.org/10.18653/v1/2024.findings-emnlp.963} {Narrative-of-thought: Improving temporal reasoning of large language models via recounted narratives}.
\newblock In \emph{Findings of the Association for Computational Linguistics: EMNLP 2024}, pages 16507--16530, Miami, Florida, USA. Association for Computational Linguistics.

\bibitem[{Zhang et~al.(2024{\natexlab{d}})Zhang, Blum, Choji, Shah, and Vempala}]{zhang-etal-2024-ultra}
Xinliang~Frederick Zhang, Carter Blum, Temma Choji, Shalin Shah, and Alakananda Vempala. 2024{\natexlab{d}}.
\newblock \href {https://doi.org/10.18653/v1/2024.findings-acl.487} {{ULTRA}: Unleash {LLM}s' potential for event argument extraction through hierarchical modeling and pair-wise self-refinement}.
\newblock In \emph{Findings of the Association for Computational Linguistics: ACL 2024}, pages 8172--8185, Bangkok, Thailand. Association for Computational Linguistics.

\bibitem[{Zhang et~al.(2024{\natexlab{e}})Zhang, Bo, Ma, Li, Chen, Dai, Zhu, Dong, and Wen}]{zhang2024survey}
Zeyu Zhang, Xiaohe Bo, Chen Ma, Rui Li, Xu~Chen, Quanyu Dai, Jieming Zhu, Zhenhua Dong, and Ji-Rong Wen. 2024{\natexlab{e}}.
\newblock A survey on the memory mechanism of large language model based agents.
\newblock \emph{arXiv preprint arXiv:2404.13501}.

\bibitem[{Zhong et~al.(2024)Zhong, Guo, Gao, Ye, and Wang}]{zhong2024memorybank}
Wanjun Zhong, Lianghong Guo, Qiqi Gao, He~Ye, and Yanlin Wang. 2024.
\newblock Memorybank: Enhancing large language models with long-term memory.
\newblock In \emph{Proceedings of the AAAI Conference on Artificial Intelligence}, volume~38, pages 19724--19731.

\end{thebibliography}
